\documentclass{article}

\usepackage{arxiv}
\usepackage{amsmath}

\usepackage[utf8]{inputenc} 
\usepackage[T1]{fontenc}    
\usepackage{hyperref}       
\usepackage{url}            
\usepackage{booktabs}       
\usepackage{amsfonts}       
\usepackage{nicefrac}       
\usepackage{microtype}      
\usepackage{cleveref}       
\usepackage{lipsum}         
\usepackage{graphicx}
\usepackage[square,sort,comma,numbers]{natbib}
\usepackage{doi}
\usepackage{textcomp}
\usepackage{booktabs}
\usepackage{subcaption}
\usepackage{xcolor}

\title{
ConKeD++ - Improving descriptor learning for retinal image registration: A comprehensive study of contrastive losses
}

\newif\ifuniqueAffiliation
\uniqueAffiliationtrue

\ifuniqueAffiliation 

\usepackage{authblk}

\newbox{\orcid}\sbox{\orcid}{\includegraphics[scale=0.06]{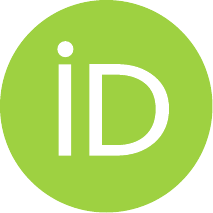}} 
\author[1,2]{%
	\href{https://orcid.org/0000-0001-7824-8098}{\usebox{\orcid}\hspace{1mm}David Rivas-Villar\thanks{\texttt{Corresponding author: david.rivas.villar@udc.es}}}%
}
\author[1,2]{%
	\href{https://orcid.org/0000-0002-9080-9836}{\usebox{\orcid}\hspace{1mm}Álvaro S. Hervella\thanks{\texttt{a.suarezh@udc.es}}}%
}
\author[1,2]{%
	\href{https://orcid.org/0000-0003-4407-9091}{\usebox{\orcid}\hspace{1mm}José Rouco \thanks{\texttt{jrouco@udc.es}}}%
}
\author[1,2]{%
	\href{https://orcid.org/0000-0002-0125-3064}{\usebox{\orcid}\hspace{1mm}Jorge Novo\thanks{\texttt{jnovo@udc.es}}}%
}
\affil[1]{Grupo VARPA, Instituto de Investigacion Biomédica de A Coru\~na (INIBIC), Universidade da Coru\~na, 15006 A Coru\~na, Spain}
\affil[2]{Departamento de Ciencias de la Computación y Tecnologías de la Información, Universidade da Coru\~na, A Coruña, 15071, A Coruña, Spain}
\fi

\renewcommand{\shorttitle}{ConKeD++}

\hypersetup{
pdftitle={
ConKeD++ - Improving descriptor learning for retinal image registration: A comprehensive study of contrastive losses
},
pdfsubject={cs.CV},
pdfauthor={David~Rivas-Villar, Álvaro S.~Hervella, José~Rouco, Jorge~Novo},
pdfkeywords={Contrastive Learning, Feature-based Registration, Image Registration, Medical Imaging, Retinal Image Registration},
}

\begin{document}
\maketitle

\begin{abstract}
Self-supervised contrastive learning has emerged as one of the most successful deep learning paradigms. In this regard, it has seen extensive use in image registration and, more recently, in the particular field of medical image registration. In this work, we propose to test and extend and improve a state-of-the-art framework for color fundus image registration, ConKeD. Using the ConKeD framework we test multiple loss functions, adapting them to the framework and the application domain. Furthermore, we evaluate our models using the standarized benchmark dataset FIRE as well as several datasets that have never been used before for color fundus registration, for which we are releasing the pairing data as well as a standardized evaluation approach. Our work demonstrates state-of-the-art performance across all datasets and metrics demonstrating several advantages over current SOTA color fundus registration methods.
\end{abstract}


\section{Introduction}
\label{sec:intro}


Image registration, also known as image alignment, is the process in which two images are spatially transformed to achieve a pixel-wise matching between their contents. In practice, one of the images is typically used as reference, the fixed image, while the other, the moving image, is the one being transformed to match the fixed one. The images to register are usually of the same subject but are taken from different view points and/or at different times. Therefore, the content depicted in the images may vary due to changes in the subject (resulting from the passage of time, disease, etc.) or capture conditions (such as perspective, lighting, etc.), among others. 


Registering medical images is highly important in modern clinical practice as it has multiple uses \cite{survey}. Firstly, image alignment facilitates the simultaneous analysis of multiple images, allowing clinicians to draw better conclusions \cite{book_mir}. However, manual alignment is not feasible as part of day-to-day clinical workflows. Secondly, image registration is an important component of  numerous computer-aided diagnosis systems \cite{morita,yanase}. Therefore, the development of robust and efficient registration methods for medical imaging is  highly desirable. However, the development of these methods is complex due to the particularities and challenges of each imaging modality and the potential changes in the underlying anatomical structures caused by the presence of disease. The latter aspect is particularly relevant, as one of the primary applications of medical image registration is in longitudinal studies \cite{Narasimha} in which images from different time frames are compared. Therefore, automatic methods should be robust against disease progression or remission. 

Within medical image registration, Retinal Image Registration (RIR), i.e. alignment images from the eyes, is very relevant. The eyes are the only organs that allow for non-invasive \textit{in vivo} observation of the blood vessels and neuronal tissue \cite{forrester2020eye}. There are multiple imaging modalities created with the goal of capturing images of the eye fundus and underlying tissue, i.e. the retina. Specifically,  color fundus (CF) images are widely used as they are non-invasive and cost-effective due the low price of the capture devices. Additionally, they have shown to be highly effective for the diagnosis of multiple diseases \cite{costeffec,Kanski}. However, CF images have certain particularities that complicate the registration process. These images are, essentially, photographies captured with a camera. This can lead to multiple imperfections due to, for instance, sub-optimal placement of the camera, movements by the patient, sub-optimal focus or lighting settings, etc. Furthermore, they also present features that are unique to retinal images, like previously described. These, coupled with the possible imaging imperfections and the characteristic variations of CF images, make the registration process a highly challenging task, requiring robust and adaptable methods.

Commonly, automatic registration methods are divided into three groups: Feature Based Registration (FBR), which is based on the use of discrete keypoints; Intensity Based Registration (IBR), which is based on the direct comparison of the intensity values of the image; and Direct Parameter Regression (DPR), which directly estimates a deformation field or transformation matrix from the input images. 
Broadly, in medical imaging most of the current state-of-the-art methods are IBR or DPR \cite{voxel,dlir,Haonan}.  IBR methods are based on the notion of maximizing a similarity metric among images by altering  the spatial transformation applied to the moving image. There are multiple available similarity metrics and the choice to use one or another depends on the modalities of the input images \cite{Pluim,Balakrishnan}. Additionally, these metrics can also be computed using deep learning \cite{cheng18}. On the other hand, DPR methods directly predict deformation fields from the input images \cite{Haskins,voxel}. In order to do so, a deep neural network is trained so that its output, a deformation field, is applied to the moving image. Then a similarity metric is computed between the fixed and the transformed moving image and used as the loss function. Thus, in order to minimize this loss function, the network must accurately deform the moving image to match the fixed image.
It is worth noting that, whereas IBR methods iteratively maximize a similarity metric to achieve the final registration, DPR methods use the similarity metric to guide the training of a neural network. This network learns to predict suitable transformations from the input data, hence not requiring any optimization of the similarity metric at inference time.

Due to the particular features of RIR, most methods in this domain are FBR \cite{rivas3, eccv20, rivas-vienna, Yiqian22}. The most relevant patterns to register in retinal images are the blood vessel tree and the optic disc. However, these occupy a relatively small portion of the images, which are generally filled with an uniform background. The small size and limited number of these structures complicate the adaptation of some medical registration methods to RIR. For instance, DPR methods tend to overfit the deformation fields to these structures while ignoring the rest of the image. Furthermore, retinal images and CF images in particular, can present large displacements and low overlapping between images in a pair. The photographic nature of CF images (and other retinal modalities) can also cause additional intra-pair differences. These characteristics are detrimental to IBR methods as a robust similarity metric is difficult to obtain under these conditions. Thus, in RIR, most methods are FBR. These methods are based on keypoints, which are distinctive spatial locations that can be repeatably detected in the images to register. These keypoints are used to drive the registration process by estimating the transformation based on matches between keypoints in the fixed and moving images. Usually, to aid in calculating the transformation, keypoints are coupled with descriptors. Descriptors are feature representations, that, ideally, uniquely characterize each keypoint, allowing for fast and easy distinction. The keypoints themselves can be generic (i.e., valid for any image) or domain specific (i.e., only valid for a particular type of images). Finally, as keypoints are easily displayable, FBR methods are inherently explainable.





Additionally, registration methods can also be separated between deep learning and classical methods. Classical methods are still widely used. However, deep learning approaches have advantages over classical methods which make them more desirable. End-to-end training is one of such advantages as it eliminates the need for feature engineering. Furthermore, deep learning-based approaches are usually more flexible as they can be easily adapted to the input data following their data-driven learning process. Moreover, the data-driven training can also make them more robust to changes in capture conditions such as illumination or focus. 


The registration of CF images is currently moving away from the dominance of classical approaches and towards deep learning methods. Deep learning methods are desirable, as previously stated, due to their increase adaptability and flexibility. This is specially relevant in RIR and, particularly, in CF registration due to the multiple sources of variability in this imaging modality. The registration performance in CF images is solely evaluated  using the benchmark dataset FIRE \cite{fire} as the reference. This reliance on a single dataset may bias these approaches towards achieving optimal performance on FIRE while compromising their generalization to other datasets and, thus, limiting their real-world applicability. In this dataset, the best results are obtained by the classical method VOTUS \cite{votus}. However, when considering the different categories of the dataset in isolation, we can see that the best classical approaches, such as VOTUS or REMPE \cite{rempe}, produce worse results than current deep learning approaches \cite{eccv20, rivas3} in the category with pathological progression. This category is the most relevant for medical practice. This highlights the added adaptability of deep learning approaches.

In the case of VOTUS, this method obtains especially good results in the category with the lowest overlapping among images in each pair. This can be explained by the use of a  quadratic transformation which has additional degrees of freedom (compared to the rest of the state of the art methods). This adds extra complexity to their optimization process. It is worth noting that the category with low overlapping is more prevalent, comprising a larger number of images compared to the category with disease progression. As a result, it significantly influences the overall dataset score, thus explaining why VOTUS achieves the best overall results. Meanwhile, REMPE, and particularly deep learning methods such as SuperRetina \cite{eccv20} and ConKeD \cite{rivas3}, are burdened by their worse performance in category P, which affects their overall score more than the performance in A, as P has more images. REMPE uses an eye-specific transformation while SuperRetina and ConKeD employ the homographic transformation. Finally, other relevant deep learning methods are the previous works of Rivas-Villar et al. \cite{rivas, rivas2} although they do not match the results of current deep learning methods.


Regarding the best performing classical methods, VOTUS \cite{votus} registers images by creating graphs for the arterio-venous tree. These graphs are then matched using a novel algorithm (Vessel Optimal Transform) that relies on classical image features. The transformation matrix is computed using DeSAC (Deterministic Sample And Consensus). On the other hand, REMPE \cite{rempe} uses discrete keypoints (SIFT and blood vessel bifurcations). Then, using RANSAC (Random Sample Consensus) \cite{ransac} and PSO (Particle Swarm Optimization), the optimal transformation is computed according to an eye-specific model.

Regarding the best performing deep learning methods, SuperRetina \cite{eccv20} adapted the natural image registration method  SuperPoint \cite{superpoint} to CF registration. They jointly train a detector and a descriptor using a ground truth of vessel keypoints. The descriptors are learned using a triplet approach. However, SuperRetina requires ample ad-hoc image pre-processing and a double inference step, which nullifies some of the advantages of deep learning methods over classical approaches. On the contrary, ConKeD does not need any of these extra steps, however, it produces slightly worse results. ConKeD also uses a ground truth of keypoints (blood vessels and crossovers) but proposes a novel way of learning descriptors, a multi-positive multi-negative metric learning approach, which improves the results from the commonly used triplet approach. In this regard, ConKeD stands out as the state of the art method in terms of descriptor learning for retinal image registration. However, it could be limited by design decisions such as the loss function.  Nevertheless, in fields related to image registration, such as image retrieval, alternative loss functions have been tested showing accurate results. Thus, these novel losses have the potential to also improve descriptor learning in our domain.


In this work, we propose to improve the ConKeD framework, and use it as a foundation to improve CF registration. Currently, ConKeD \cite{rivas3} has multiple advantages over the rest of the state of the art methods although its performance might be limited by its design decisions. Thus, we propose to address it by testing multiple contrastive learning loss functions and, more importantly, losses with different structures and properties so that the maximum amount of information can be effectively extracted from the training images. "Furthermore, since current color fundus registration evaluation is based solely on the FIRE dataset, we propose  significantly expand the evaluations. 
We incorporate two additional datasets with a larger number of registration pairs (3141 and 990, respectively, compared to the 134 pairs in the case of FIRE). These datasets offer diverse desirable features, including multiple disease grades and varied overlapping amounts.

\section{Materials and Methods}
\begin{figure*}
    \centering
    \includegraphics[width=\textwidth]{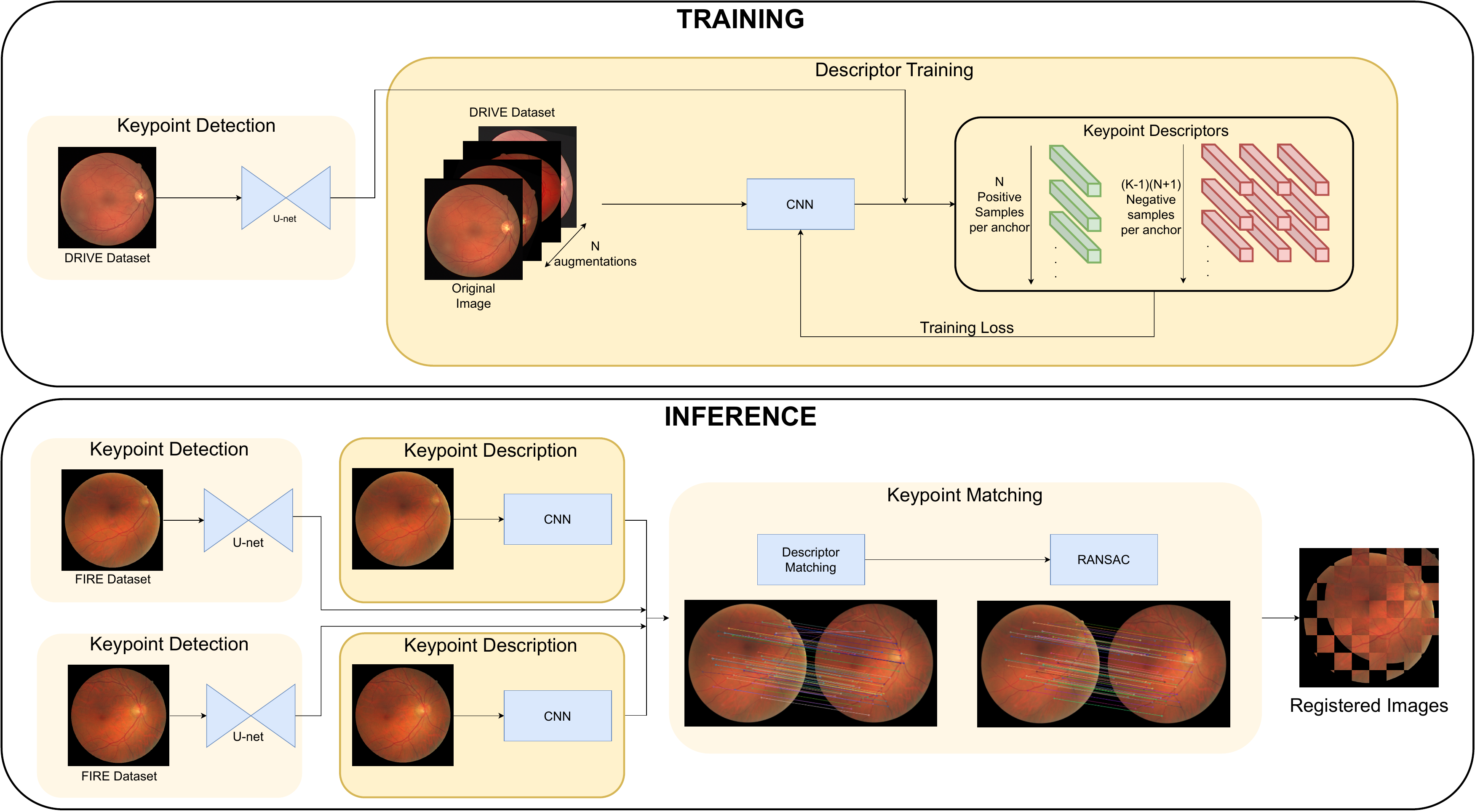}
    \caption{Overview of the method for both training and inference.}
    \label{fig:over}
\end{figure*}

In order to explore the different training losses, we use the ConKeD \cite{rivas3} framework in conjunction with the whole registration pipeline proposed in \cite{rivas3}. The complete methodology is described in Fig. \ref{fig:over}. This methodology is based around two deep neural networks. Firstly, keypoints (blood vessel crossovers and bifurcations) are detected using a network trained with a ground truth. Next, these points are used to compare pixel-wise descriptor samples in the ConKeD framework. ConKeD creates a multiviewed batch using an original image from the dataset and $N$ views or augmentations. This way, as each keypoint is selected as anchor (i.e. the point used as reference to calculate the loss function), it will have  $N$ positive samples (the same keypoint under different views) and many negatives (all the remainder keypoints are considered negatives, including the points in the anchor's own image). Using all these samples, in combination with any suitable loss, the network is trained, learning to match positive samples across views while differentiating them from the negative examples. At inference, the keypoints are detected by the first network, described by the second one, matched and, finally, from the matching keypoints, RANSAC creates a projective transform to align both images in each registration pair.

\subsection{Baseline methodology}

\subsubsection{Keypoint detection}

The first step of the method is the detection of the keypoints, which in this case are blood vessel crossovers and bifurcations. These keypoints are domain specific (i.e., particular from the Retinal image domain), very spatially precise (i.e. can be finely detected) and distinctive, which allows for an accurate matching. However, the number of blood vessel crossovers and bifurcations varies from eye to eye and from image to image. This is due to the differences in the underlying retinal morphology, the differences in the capture viewpoint, and the possible progression or remission of pathologies. 

To detect these keypoints we train a CNN to generate heatmaps with peaks at the locations where the keypoints are present \cite{alvaro_cmpb,rivas3}. Using heatmaps instead of a binary maps helps the network to learn in spite of the high class imbalance between the keypoints and the background. If the ground truth labels were binary, the number of positive instances (i.e. the keypoints) would be drastically lower than the number of instances belonging to the negative class (any other pixel in the image). The heatmaps peak at the exact location of each keypoint, with values diminishing progressively as the distance from each keypoint increases. Overall, the heatmaps increase the information made available to the network during training  which, in turn, improves the learning process \cite{alvaro_cmpb, rivas3}. 

In order to separate crossovers from bifurcations the network generates two separate heatmaps. Furthermore, in order to incentivize the network to detect all the possible keypoints, even if it cannot reliably tell if they are bifurcations or crossovers, a third heatmap is also generated, containing both crossovers and bifurcations. The network is trained using the mean squared error (MSE) between the predicted heatmaps and the ground truth heatmaps as the loss function. To create the ground truth heatmaps the ground truth binary maps are convolved using a Gaussian kernel.

Finally, at inference time, the discrete keypoints are extracted from the predicted  heatmaps by using a local maxima filter with an intensity threshold \cite{alvaro_cmpb, rivas3}.

\subsubsection{Keypoint description}

The second step is the keypoint description. This is accomplished using a CNN that generates a dense descriptor block, meaning a descriptor is produced for every pixel in the input image. In order to train this network, we use ConKeD. This framework follows a novel multi-positive multi-negative contrastive learning approach. Previously, descriptors for image registration have been usually learned using triplet-based approaches. These approaches pair each anchor point with a single positive sample and a negative one. Therefore, for each anchor point, only two comparisons are performed. However, ConKeD leverages all the available keypoints in its multiviewed batches, creating multiple comparisons with several positive and negative samples for each anchor point. The quantitative advantages of using this method were already proven in \cite{rivas3} showing that this method facilitates the training by increasing the amount of contrastive comparisions. Furthermore, using a multi-positive and multi-negative approach there is no need for explicit mining \cite{fastap}.

The network is trained employing a multiviewed batch. This batch contains a single original image from the training set. From this image, $N$ different augmentations (views) are generated using both color and spatial transformations. Therefore, the total size of this batch is $1+N$. As the network creates a dense block of descriptors, each pixel in each image of the batch will have a suitable descriptor. If there are $K$ keypoints per image on average, the multiviewed batch will have a total of $K(1+N)$ keypoints. Each anchor point will have $N$ positive samples, each one of them belonging to a different view (i.e. image augmentation) from the total $N$ generated views included in the multiviewed batch. Likewise, each anchor point will also have $(K-1)(1+N)$ negative samples, which include every other keypoint that is not a positive sample (this also includes the other keypoints from the same view). Additionally, it is worth noting that all the negative samples are hard-negatives (i.e. closely related and similar to the anchor point). This is because all the keypoints have been detected by the network, hence they are either blood vessel crossovers or bifurcations located in the arterio-venous tree.  Using hard negatives is beneficial since the network will have to learn to create accurate descriptors capable of distinguishing these largely similar points which would result in better matching performance and more accurate registration.

In this work we propose to explore several loss functions, which are described in depth in section \ref{ssec:losses}. To create the contrastive comparisons in those losses, we compare the descriptors of the anchor points with their positive and negative samples. To do so, we always use cosine similarity as the metric across all the used loss functions. Similarly to other works \cite{simclr, vinyals, supcon}, we apply L2 normalization to the output of the neural network, which, combined with the dot product used in each of the losses, creates a metric equivalent to the cosine similarity. Moreover, this normalization helps contrastive losses to perform intrinsic hard positive/negative mining \cite{supcon}. 

\subsubsection{Matching and transformation computation}

Once the keypoints are detected and described they can be used to estimate the transformation that aligns the images. Firstly, the appropriate pixel descriptors are selected from the dense map (same size as the input image) produced by the network. These pixel descriptors correspond to the locations of keypoints detected by the detection network (i.e., crossovers and bifurcations). Then, these descriptors are matched and compared using the cosine similarity between the two images that make up the registration pair. This results in a set of paired keypoints that includes all the keypoints for which an appropriate match was found. Using this  set of paired keypoints we can estimate the transformation that registers the images with the well-known RANSAC algorithm \cite{ransac}. RANSAC is able to separate the data in inliers and outliers (in the case that some were present even after descriptor matching) and only use the inliers to compute the transformation. The resulting transformation matrix aligns the input images, completing the registration pipeline.


\subsection{Losses}
\label{ssec:losses}

In this paper, we explore different loss functions for the ConKeD framework. The aim is to improve the results previously obtained and increase the data efficiency of the proposed method. In particular we study the following losses: 

\subsubsection{SupCon Loss}

SupCon Loss \cite{supcon} is a loss function that was originally designed for multi-positive multi-negative contrastive learning, but in the context of supervised image classification. In the case of the ConKeD framework, we can define this loss using the following notation. Starting from a multiviewed batch of $N+1$ images, each of them containing $K$ keypoints, we define $S$ as the set of all the keypoints such that its cardinality is $|S| = (N+1)K$. Furthermore, $s \in S$ is an arbitrary anchor keypoint taken from $S$. Let $P(s)$ be the set of the corresponding positive samples for $s$ (i.e. other views of the anchor keypoint) and $p \in P(s)$ be one of these positive samples. Additionally,  $A(s) = S \setminus {s}$ is the set of all keypoints minus the anchor (i.e. this set contains both the positive and the negative samples for $s$). Considering this, the SupCon loss function for ConKeD is defined as:

\begin{equation}
    \mathcal{L}_{sup} = \sum_{s \in S}{\frac{-1}{|P(s)|} \sum_{p \in P(s)}{\log \frac{exp(z_{s} \cdot z_{p} / \tau)}{  \sum\limits_{a \in A(s)}{exp(z_{s} \cdot z_{a}/\tau)}}},}
\end{equation}
where $z$ represents the network output for any keypoint and symbol $\cdot$ to denotes the dot product. Additionally, $\tau \in R^+$ is the scalar temperature parameter.

\subsubsection{MP-InfoNCE Loss}
InfoNCE \cite{vinyals} is a loss function that is commonly used across multiple self-supervised learning approaches in many domains \cite{vinyals,moco,simclr}. Originally, this loss function was only suitable for single-positive multi-negative contrastive learning. However, in \cite{rivas3}, it was adapted to the multi-positive multi-negative ConKeD framework. This adapted version, know as MP-InfoNCE, is computed as follows:

\begin{equation}
    \mathcal{L}_{NCE} =\sum_{i \in I}{\frac{1}{|I'|}\sum_{x_{i} \in X_{i}}{\frac{1}{|X_{i}|}\sum_{\substack{j \in I \\ j>i}}{-\log \frac{exp(z_{x_{i}} \cdot z_{p_j(x_{i})})/ \tau)}{\sum\limits_{c \in C_{i,j}(x_{i})}{exp(z_{x_{i}} \cdot z_{c})/\tau}}}}},
\end{equation}

where $I$ is the set of images which make up the multiviewed batch, such that its cardinality is $|I| = (N+1)$.  $I'$ denotes the set of all the subsets of two elements of $I$ such that its cardinality is  $|I'| = {I \choose 2}$. We use $i \in I$ and $j \in I$ with $j>i$ to denote the indexes of two different images from $I$ forming a pair in $I'$. $X_{i}$ and $X_{j}$ are the sets composed by all of the keypoints present in images $i$ and $j$, respectively. Then, $x_{i}$ is an arbitrary anchor keypoint from $X_{i}$ and $p_j(x_{i})$ is its corresponding positive sample in $X_{j}$. Additionally, $C_{i,j}(x) = (X_{i} \cup X_{j}) \setminus x_{i} $ is the set of all the keypoints in images $i$ and $j$ minus the anchor keypoint (i.e. all the negatives and the positive sample).

\subsubsection{MP-N-Pair Loss}

N-Pair loss \cite{npair} was the first multi-negative loss proposed for contrastive learning. However, this paradigm did not start to trend upwards until after info-NCE \cite{vinyals} was introduced, which caused the original work proposing the N-Pair loss to go largely unnoticed. It should be noted that info-NCE \cite{vinyals} can be interpreted as generalization of N-Pair loss \cite{npair} as the latter can be seen as info-NCE \cite{vinyals} with a temperature parameter of $\tau = 1$. Therefore, it is also possible to define a multi-positive multi-negative version of N-Pair.  This adapted version, which we call MP-N-Pair, is computed as follows:

\begin{equation}
    \mathcal{L}_{N-Pair} =\sum_{i \in I}{\frac{1}{|I'|}\sum_{x_{i} \in X_{i}}{\frac{1}{|X_{i}|}\sum_{\substack{j \in I \\ j>i}}{-\log \frac{exp(z_{x_{i}} \cdot z_{p_j(x_{i})}))}{\sum\limits_{c \in C_{i,j}(x_{i})}{exp(z_{x} \cdot z_{c})}}}}},
\end{equation}

where $I$ is the set of images which compose our multiviewed batch such that its cardinality is $|I| = (N+1)$ . $I'$ denotes the subsets of two elements of $I$ such that its cardinality is  $|I'| = {I \choose 2}$. We define $i \in I$ and $j \in I$ such that $j>i$ are the indexes of two different images from $I$ forming a pair in $I'$. Let $X_{i}$ and $X_{j}$  be the sets of all keypoints in images $i$ and $j$. In this case, $x_{i}$ is an arbitrary anchor keypoint taken from $X_{i}$ and $p_j(x_{i})$ is its corresponding positive sample from  $X_{j}$. Additionally, $C_{i,j}(x) = (X_{i} \cup X_{j}) \setminus x_{i} $ is the set of all the keypoints in images $i$ and $j$ except for the anchor keypoint (i.e. all the negatives and the positive sample).

\subsubsection{FastAP Loss}

FastAP loss is based on the notion of minimizing the Average Precision (AP). AP has seen wide use in information retrieval and adjacent fields such as image retrieval. Feature matching can be seen as derivative field of information retrieval. The goal is to rank all positive samples above the negative ones. Furthermore, there should be an ordering between positive and negative ones as their relevance will vary depending on the anchor. Therefore, to calculate AP, we would need to create the AUC of the precision recall curve. In this case, the precision represents the ratio of relevant items retrieved (i.e. positive samples) to the total items retrieved (all samples, positives and negatives). On the other hand, the recall is the ratio of relevant items retrieved to the total relevant items in the dataset. Therefore, FastAP works differently than the three previous losses as it does not continue bringing samples closer (anchor and positives) or pushing them apart (anchor and negatives) once they are all correctly placed.

However, this loss has an inherit advantage derived from its structured (i.e., ranking-wise) nature. The loss inherently creates a ranking of negative samples, thus creating an implicitly structured embedding space, which might benefit the end goal task of matching descriptors for image registration.

In general, AP losses are challenging to optimize as it they are non-decomposable over ranked items and differentiation through the sorting operations is complex. The FastAP Loss \cite{fastap} was proposed to address these issues, creating an approximation of the classical AP that is differentiable and efficient. To do so, the authors use the relation between the AP and AUPR (Area Under the Precision-Recall curve), as when there are infinite elements both are related asymptotically \cite{fastap}. This allows precision and recall to be casted as parametric function of distance. Moreover, the normalization of the descriptors in the output of the network (using L2-norm) allows to create a bounded range of possible distances (from 0 to 2). Therefore, the distances can be interpreted as a histogram, to be quantized in a set of discrete bins. With these notions, we can define the FastAP Loss as:

\begin{equation}
    \mathcal{L}_{FastAP} =\frac{1}{|S|}\sum_{i =  1}^{S}{\frac{1}{M^+_{s_{i}}}\sum_{j= 1}^{Q}\frac{H^{+}_{j}h^{+}_{j}}{H_{j}}},
\end{equation}

where $S$ is the set of samples (i.e. keypoints) in the batch. The cardinality of $S$ is $|S|$, which in our case of a multiviewed batch of $N+1$ images, each of them containing $K$ keypoints would be $|S| =(N+1)K$. $s_{i}$ represents a particular sample from $S$, identified by its index $i$. This sample acts as the anchor and  $M^{+}_{s_{i}}$ is the number of positives that this sample has in the batch. Following the ConKeD approach $M^{+}_{s_{i}} = N$. $Q$ is the number of discrete bins in which the histogram is quantized.  $h_{j}$ is the number of samples that fall in the $j-$th bin in the histogram and $H_{j} = \sum_{k \le j} h_{k}$ is the cumulative sum of the histogram up to that bin. Likewise, $h^{+}_{j}$ is the number of positive samples of $s_{i}$ in the $j-$th bin of the histogram and $H^{+}_{j}$ is its cumulative sum.

\subsection{Experimental setup}

\subsubsection{Datasets} \label{sssec:datasets}

We train both the detection and descriptor network using DRIVE \cite{drive}. DRIVE is a public dataset composed of images captured in a diabetic retinopathy screening program using a Canon CR5 non-mydriatic 3CCD camera with a FOV of 45 degrees. The images have a resolution of $584\times565$ pixels. DRIVE is composed of 40 images and is equally divided in training and test sets, each consisting of 20 images. This dataset has a ground truth of vessel crossovers and bifurcations that we use to train our keypoint detector \cite{Abbasi}. The original labeling from DRIVE is converted to heatmaps, as previously explained. Additionally, it should be noted that these labels are only used to train the keypoint detection network, the description network does not require any ground truth labels.

The evaluation is performed on multiple datasets. First, we use FIRE \cite{fire}, which is the standard benchmark dataset for CF registration, as it is the only publicly available dataset with registration ground truth. In order to perform a more robust evaluation, we have also collected the data from two other publicly available datasets and manually established corresponding image pairs for evaluating the registration. These datasets are: Longitudinal diabetic retinopathy screening data (LongDRS) \cite{ldrs} and DeepDRiD \cite{deepdrid}. Both of these datasets include multiple images with diabetic retinopathy. In particular LongDRS includes two separate visits, each with multiple images. This allows to create low-overlapping pairs from the same visit and and inter-visit pairs with varied degrees of overlapping. On the other hand, DeepDRiD, includes images from all the severity stages of diabetic retinopathy as well as images with different types of artifacts that complicate the image registration. The pairing data used to evaluate the registration on these datasets is made available to facilitate the evaluation and comparison of future works.


The FIRE \cite{fire} dataset is composed of 129 unique retinal images from 39 patients, which form a total 134 registration pairs. These images were acquired with a Nidek AFC-210 fundus camera with a resolution of 2912x2912 pixels and a FOV of 45°. This dataset is split in three categories depending on the characteristics of each image pair. The total 134 image pairs are divided in 71 from category S, 49 from P and 14 from A. Category S has high overlapping between the images of the registration pair. On the other hand, category P has low overlapping. Finally, category A has high overlapping but the images composing each pair show pathology progression between one another. This makes  category A highly complex but also highly relevant for clinical practice. FIRE has a registration ground truth consisting of labeled keypoints. These are used to compute the distance between the optimal and the estimated registration. Finally, it should be noted that one of the images, from the P category, presents an incorrectly labeled keypoint. In this case, we opted for discarding the incorrectly labeled keypoint and evaluate that particular image with one less control point.

LongDRS is a dataset containing images from a diabetic retinopathy screening program. In particular it contains the images of 70 patients. Each patient had four images per eye taken during two separate visits (with a one-year interval), resulting in a total of eight images per eye and sixteen images for each patient. Therefore, there are 1120 images in the LongDRS dataset. The images are captured using a Topcon TRC-NW6S nonmydriatic digital funds camera with a 45º field-of-view. The fundus images are $2000 \times 1312$ pixels in size. This dataset does have a pseudo-ground truth for image mosaicking created by two different automated methods. However, it is not suitable for pair-based registration and, as it depends on automated methods, it is not reliable as ground truth. In our case, the possible registration pairs are created by an expert by observing the available images and deciding which ones overlap in sufficient area so they can be assigned unequivocally to the same eye. This dataset allows for different types of evaluations due to the images from two different visits. Firstly, the images can be registered in a intra-visit manner. These image pairs normally have mid-to-very-low overlapping (as they are created for mosaicking) but, given that they are captured at the same visit, they show no disease progression or morphology changes of any kind. Secondly, the images can also be registered in a inter-visit manner. These image pairs offer  different degrees of overlapping, ranging from very high to very low overlapping. Therefore, this set is particularly interesting as it combines numerous overlapping settings with a time separation between visits. This evaluation variant is entirely absent from the evaluation conducted on FIRE, which features either time separation or low overlapping, but not both simultaneously. Overall, the 1120 images composing the dataset were manually checked to find suitable registration pairs. This way, we build a registration dataset with a total of 3141 image pairs can that can be reliably recovered. Out of these, 1839 are inter-visit and 1302 are intra-visit (647 from the first visit and 655 from the second one).

DeepDRiD \cite{deepdrid} is a dataset originally created for diabetic retinopathy assessment, grading and diagnosis. It contains 2,000 fundus images with diabetic retinopathy from 500 patients and 256 ultra-widefield images from 128 patients. It should be noted that we only use the fundus images, not the ultra-wide ones. The fundus images contain samples from all grades of severity of diabetic retinopathy, making this dataset very representative as it contains images from patients who are largely healthy as well as patients who suffer from severe diabetic retinopathy. Furthermore, this dataset also presents multiple images with artifacts, clarity issues and, overall, image quality issues. While these may impede accurate grading of diabetic retinopathy, they do not impede manual registration (i.e., pairs are technically registrable), which makes them desirable to test the automatic method’s robustness. Therefore this dataset is very relevant as a test of robustness for automatic approaches. For each patient, there are images from both left and right eyes as well as from two different viewpoints (macula-centered and optic-disc-centered images). Using these two separate viewpoints we can construct the registration pairs, i.e. aligning these two corresponding images for each eye. Given that there are 500 patients, all of them with images from both eyes, there is a total of 1000 potential registration pairs. From these, we removed 10 pairs as they were not suitable (i.e., the same image was repeated twice, the images were already registered and overlapped, etc.). Thus, we build a pairwise registration dataset containing a total of 990 registrable image pairs. Although these images are categorized as macula- and optic-disc-centered there is a high degree of variation of the centers of the images. Therefore, the pairs oscillate between high and low overlapping depending on this factor.

Finally, it should be noted that we are releasing the files corresponding to the image registration pairings created for  both LongDRS and DeepDRiD in order to facilitate evaluation and comparison for future works \footnote{ http://varpa.org/research/ophtalmology.html\#Registration}.

\subsubsection{Keypoint detection}

The training of the keypoint detection network is performed as in the state of the art \cite{alvaro_cmpb, rivas3}. We use the training set of the DRIVE dataset \cite{drive}, that is, 20 images. However, from these 20 available images, we reserve 20\% for validations, thus this network is trained with 15 images in total.  We use an  U-Net \cite{ronneberger15} network architecture as in \cite{alvaro_cmpb, rivas3}. This network is trained from scratch using Adam \cite{adam} as the optimizer, employing learning rate decay. The learning rate is originally set to $1e-4$. Using a a patience of 2500 batches, this learning rate is reduced by a factor of 0.1 each time, until the training stops when learning rate reaches  $1e-7$. To train the network, we use spatial and color augmentations. The spatial data augmentations consists of affine transformations with random rotations of $\pm 90^{\circ}$, random scaling of $0.9-1.1 \times imageSize$ and random shearing of $\pm 20^{\circ}$. The color augmentation consists on randomly changing image components in the HSV color
space \cite{Liskowski}. 

The value of the intensity threshold used to select keypoints at inference time is found using the test of the DRIVE dataset. The chosen value, 0.35, provided best F1-Score in this set \cite{alvaro_cmpb}.

\subsubsection{Keypoint description}

To train the description network we use the whole DRIVE dataset, that is, the test and the training sets, for a total of 40 training images. This means that this network is also trained using the DRIVE dataset image resolution. 

As network architecture we use the same one as in previous works \cite{rivas3}, a modified L2-net \cite{l2net} proposed in \cite{r2d2}. This network architecture produces a dense descriptor map with the size of the input images, thus there is a descriptor per pixel. The network is trained as in \cite{rivas3}, from scratch for 1500 epochs using Adam \cite{adam} as the optimizer with a fixed learning rate of $1e-4$. For SupCon Loss and MP-InfoNCE we fixed the temperature parameter $\tau =  0.1$ as in \cite{rivas3}. For FastAP we set the number of bins in the histogram to $Q=10$, as suggested by the original authors \cite{fastap}.

We use the same augmentation regime as in \cite{rivas3}, both for spatial and color augmentations. This means random affine transformations with rotations of $\pm 60^{\circ}$, translations of $0.25\times imageSize$ in each axis, scaling between $0.75-1.25 \times imageSize$ and shearing of $\pm 30^{\circ}$. This is also coupled with color augmentation in the form of random changes in the HSV colorspace \cite{hsv} as well as random Gaussian Noise with a mean of 0 and a standard deviation of 0.05. It should be noted that the spatial and color augmentations are applied to every augmented image while the random Gaussian noise is applied with a probability of 0.25.

It should be noted that, the descriptor training step does not use any ground truth for keypoints. Instead, we use the output of the keypoint detection network. Moreover, in the training process, the keypoints are only detected in the original images. This means that the geometric augmentations are also applied to the keypoints as well as the images themselves. In terms of the size of the multiviewed batch, we use the optimal size of 1+9 images found in \cite{rivas3}.

\subsubsection{Keypoint matching}

As previously described, the descriptors corresponding to the keypoints (i.e., crossovers and bifurcations) are matched between both images of each registration pair. This is done in the same manner as in \cite{rivas3}. In particular, this entails that, for a pair of descriptors to be considered as matching, this match should be bidirectional. This means that, in order for descriptors A and B to be a match, A has to be the closets descriptor to B and B the closest descriptor to A. The distance between descriptors for this matching is measured using cosine similarity as the metric.  Moreover, as in \cite{rivas3}, we only match bifurcations to bifurcations and crossovers to crossovers, which leverages the keypoint classification capabilities of the detection network to reduce the amount of computation in the descriptor matching.

After the description matching step, the paired keypoints are scaled from the DRIVE dataset resolution (i.e., the training resolution) to the native resolution of the test dataset (FIRE, LongDRS, or DeepDRiD). This allows for a comparable evaluation with previous state of the art works \cite{fire,rempe,rivas3}. In order to transform the moving images we use a projective transformation. This transformation, also known as homography, has been widely used in previous state-of-the-art works \cite{rivas2,eccv20,rivas3}

\subsection{Evaluation methodology}

\subsubsection{FIRE dataset}

In order to evaluate the performance of our approach on the FIRE dataset we use the metric proposed by the authors of the dataset \cite{fire}. This metric, known as Registration Score, is based on the Euclidean distance calculated between the control points of the fixed and moving images. The average distance across all the control points is used as proxy for the registration error of that image pair. Then, the obtained value can be compared to a given threshold.  If the error is lower than the threshold the registration is considered to be successful, if it is bigger it is classified as unsuccessful. The registration score metric is based on an AUC (Area Under Curve). This graph is obtained by plotting the ratio of successful to unsuccessful registrations on the Y axis, and the variable error threshold on the X axis. From this graph we can compute an AUC which is the final registration score. This way, the FIRE Registration Score is the AUC computed up to 25 pixels of error. This metric is robust to outliers and offers a reliable way to compare methods. In that regard, registration score is commonly computed for the whole dataset and in per-category manner \cite{rivas,rempe, votus, rivas3}. However, not all the works in the state of the art provide this standardized metric for the whole dataset. Therefore, in order to facilitate the comparison among all the existing works, we also compute the average (Avg.) and  weighted average (W. Avg.) among the three categories in order to measure the registration performance on the whole dataset.

Moreover we also use the VTKRS (Variable Top Keypoint Registration Score) metric proposed in \cite{rivas3} to evaluate the performance of our method. This metric takes into account the computational efficiency of the transformation estimation by computing the registration score at different amounts of keypoints. The standard registration score encourages a brute force approach to detection and description as it does not take into account the efficiency of the methods. Therefore, a simple way to boost the results of this metric would be to simply detect more points, expecting that they contain a combination that allows for a lower error registration. In contrast, by controlling the number of keypoints used in the transformation estimation, we can plot a curve that shows how well the method performs when different amounts of keypoints are allowed. Therefore, this curve provides a more complete evaluation as it takes into account the precision of both the detector and the descriptor, which are more relevant in keypoint-constrained scenarios, as opposed to the brute-forcing encouraged by the standard registration score. 
The selection of keypoints to include in the evaluation is based on their matching distance. Furthermore, as we use two separate groups of keypoints (bifurcations and crossovers), we follow the original approach \cite{rivas3} including the top-matches of each class, from top-3 to top-25.

\subsubsection{LongDRS and DeepDRiD}

Contrary to FIRE \cite{fire}, LongDRS \cite{ldrs} and DeepDRiD \cite{deepdrid} do not have registration ground truth to enable direct evaluation of the registration performance. Therefore, in order to reliably evaluate the performance of the methods in these datasets we use surrogate metrics based on the pixel-wise similarity of the images. In order to verify the suitability of these metrics, we validate them using the ground truth of the FIRE dataset. This way, we can confirm a direct correlation between these surrogate metrics and the registration performance measured with a manually annotated ground truth. These metrics are calculated over the original dataset resolution. The proposed metrics are:

\begin{itemize}
    \item DICE: we compute the DICE Score \cite{dice} between the segmented blood vessels of the fixed and moving images. As these datasets lack blood vessel ground truth, we segmented them using a state of the art network \cite{alvaro_asoc}. This way, the DICE of the images is computed as $2 \times$ the intersection of the vessels divided by the area of the vessels trees of both images summed. 

    \item Intersection Over Union (IoU): The IoU, similarly to the DICE Score is also computed using the segmented blood vessels. In this case it is the ratio of the intersection divided by the union.

    \item Intersection Over Minimum (IoM): As the segmented blood vessels are the prediction of a network rather than a ground truth they can have imperfections. Therefore, metrics such as DICE or IoU, will underestimate the performance of the registration as, even in perfect alignment cases, the upper bound of these metrics cannot be reached as the segmentation is not identical for both images in the pair. In these cases, the intersection will always be smaller than the union (IoU) or half the sum (Dice), regardless of the quality of the image registration. Therefore, we propose to use the minimum of both vessel trees in the computation of a metric, as opposed to the intersection or the sum. Thus this metric, Intersection Over Minimum, may provide a more realistic evaluation.


    \item SSIM: The structural similarity index measure (SSIM) \cite{ssim} is commonly used as similarity metric among images. However, in our experiments the original formulation of this metric was not fully reliable. In particular, there were some cases for which the registration was erroneous or not accurate enough, and still a high SSIM value was achieved. This is in part due to the uniformity (i.e. low variance) of the background in color fundus images. This issue is caused by the structural component of SSIM, which is typically formulated as:


    \begin{equation}
        s_{x,y} = \frac{\sigma_{x,y} +c_{3}}{\sigma_{x} \sigma_{y} +c_{3}},   
    \end{equation}
        where $\sigma_{x,y}$ is the convariance of $x$ and $y$ and, likewise, $\sigma_{x}$ is the covariance of $x$ and  $\sigma_{y}$ is the covariance of $y$. $c_{3}$ is a constant to stabilize the division with weak denominators. Therefore, in cases where either $\sigma_{x}$ or $\sigma_{y}$ are very near 0 (due to the aforementioned uniform background in CF images and a possibly badly computed transformation) the result of this division will be, approximately, $\frac{c_{3}}{c_{3}} = 1$ which results in an unexpectedly high SSIM value for an erroneous transformation. Therefore, to avoid this issue, we re-formulate the structural term as:
    \begin{equation}
        s_{x,y} = \frac{\sigma_{x,y}}{\sigma_{x} \sigma_{y} +c_{4}}, 
    \end{equation}
    where $c_{4}$ is a small constant, different from $c_{3}$ which serves the same purpose, to stabilize the division with weak denominators. In this case we arbitrarily set $c_{4} = 1e-10$. Removing $c_{3}$ from the numerator causes the division to not result in 1 when either $\sigma_{x}$ or $\sigma_{y}$ are very near 0. 
    This metric is computed over the images transformed to grayscale in order to mitigate any color shifting and in a multi-scale manner, using window sizes of $11, 33, 55$ and $111$

    \item Structure Metric (SM): we also use the structure term for SSIM to evaluate the registration. This term does not take into account other types of similarity (i.e. intensity and contrast) and only focuses on the structures, like the blood vessel tree, which are  the relevant patterns to register in CF images. As explained above, we use a modified version of the standard structure metric, designed to be reliable on CF images. Similarly to SSIM, SM is also computed using the images transformed to grayscale and in a multi-scale manner, using window sizes of $11, 33, 55$ and $111$.

    \item LPIPS: We employ the commonly used LPIPS Similarity metric \cite{lpips} to evaluate the registration. It should be noted that, empirically, we found the variant calculated with VGG to have a stronger correlation with the registration performance results in FIRE than the the AlexNet version. Therefore we use the LPIPS VGG.

\end{itemize}

\begin{figure}
    \centering

        \begin{subfigure}{0.32\textwidth}
            \centering
            \includegraphics[width=\textwidth]{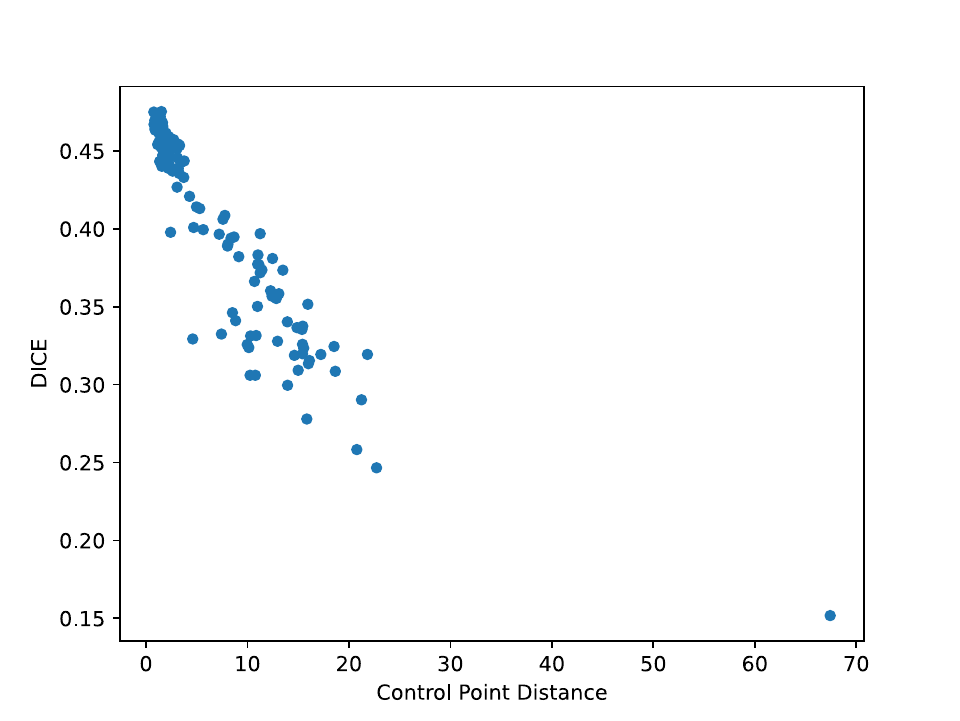}
            \caption{}
            \label{fig:corr-dice}
        \end{subfigure}
        \begin{subfigure}{0.32\textwidth}
            \centering
            \includegraphics[width=\textwidth]{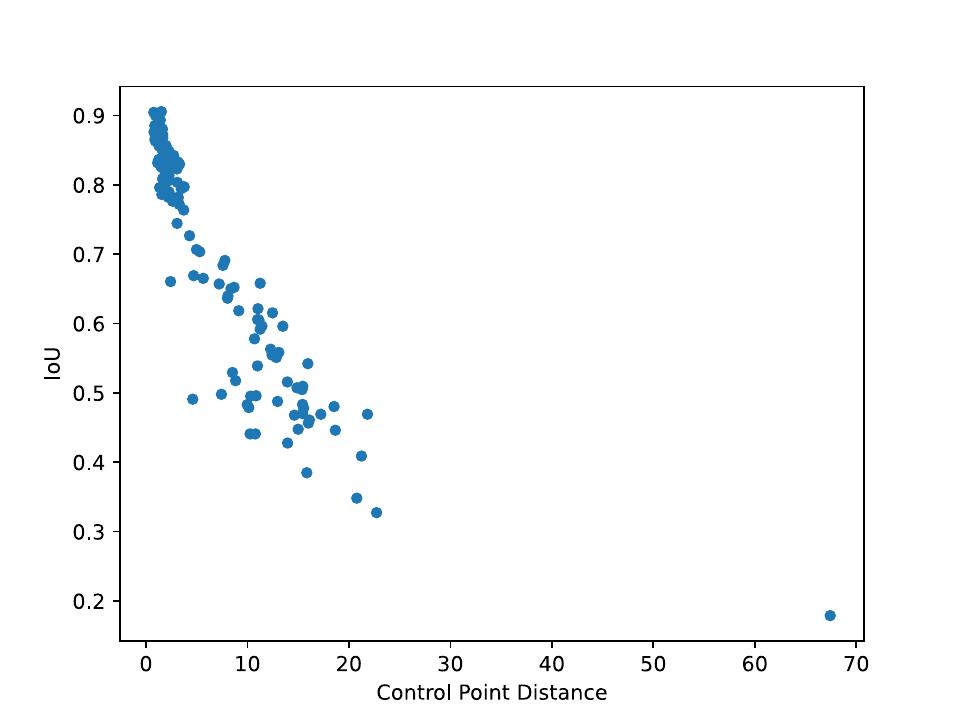}
            \caption{}
            \label{fig:corr-iou}
        \end{subfigure}
        \begin{subfigure}{0.32\textwidth}
            \centering
            \includegraphics[width=\textwidth]{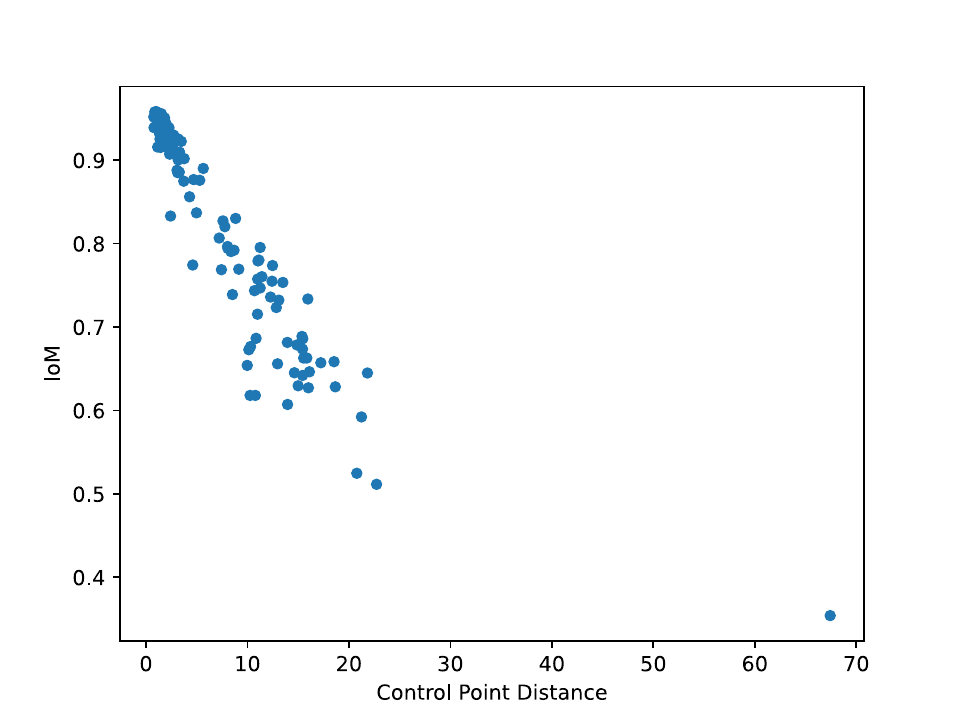}
            \caption{}
            \label{fig:corr-iom}
        \end{subfigure}
            \begin{subfigure}{0.32\textwidth}
            \centering
            \includegraphics[width=\textwidth]{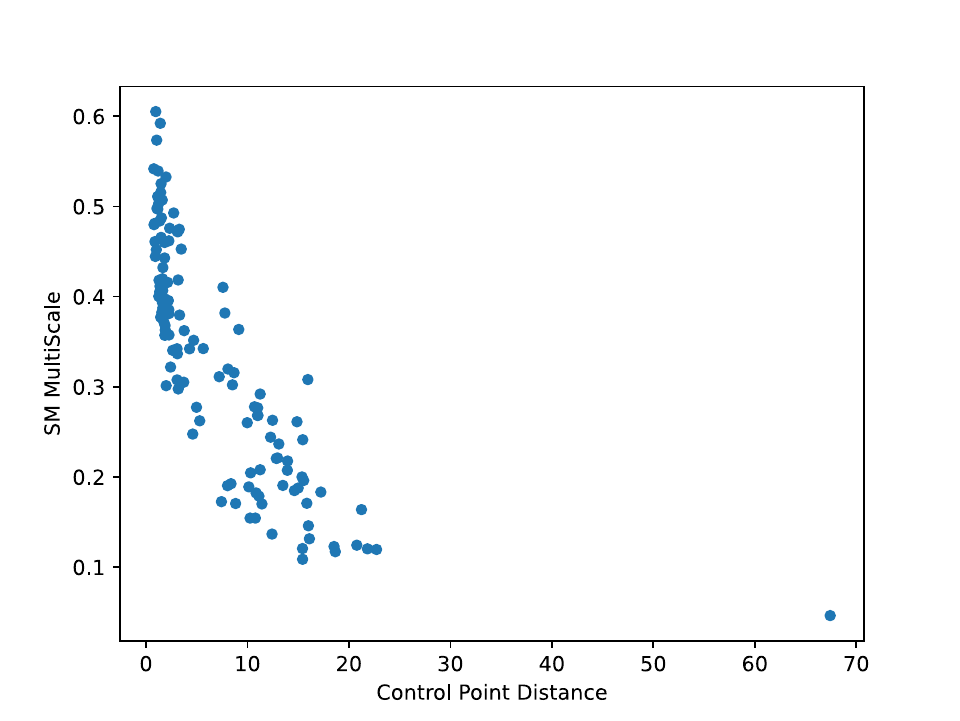}
            \caption{}
            \label{fig:_corr-sm} 
        \end{subfigure}
        \begin{subfigure}{0.32\textwidth}
            \centering
            \includegraphics[width=\textwidth]{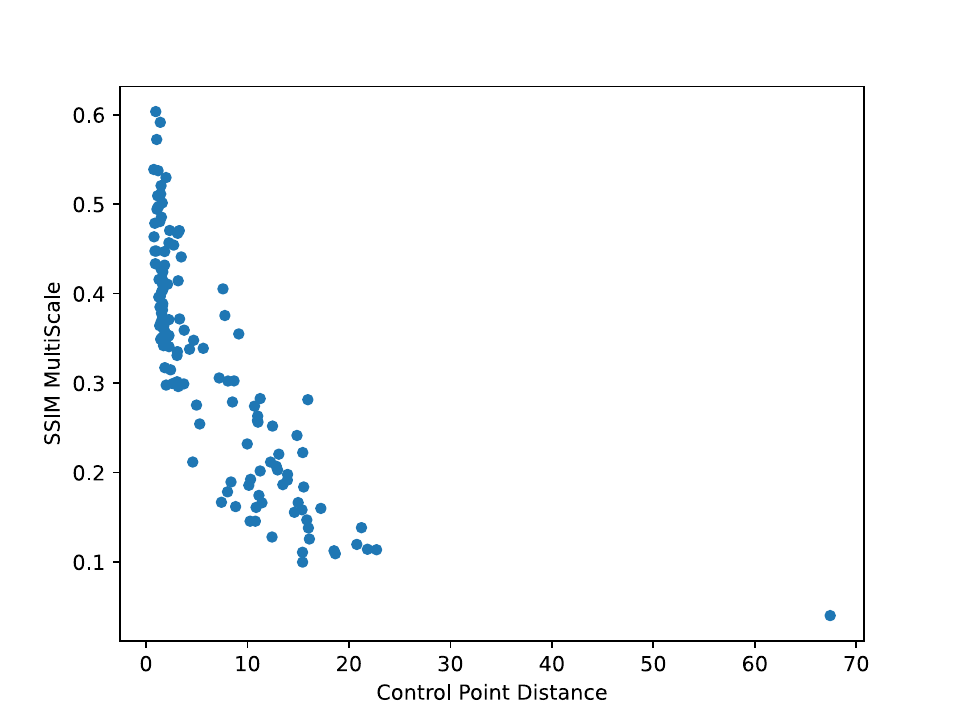}
            \caption{}
            \label{fig:corr-ssim}
        \end{subfigure}
        \begin{subfigure}{0.32\textwidth}
            \centering
            \includegraphics[width=\textwidth]{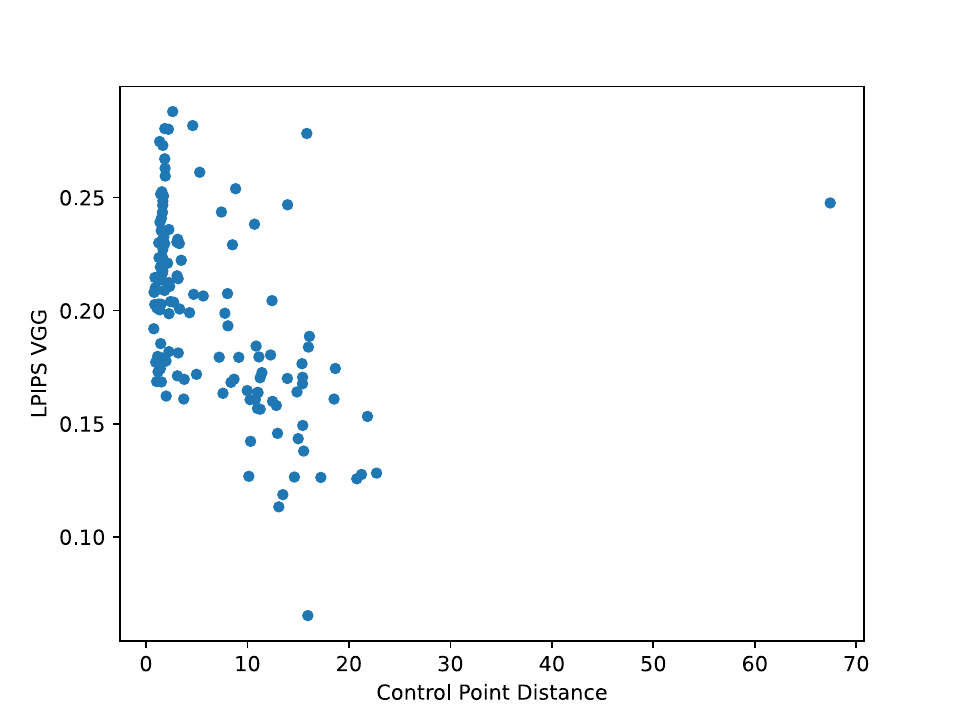}
            \caption{}
            \label{fig:_orr-lpips}
        \end{subfigure}
    
    \caption{Correlation between the proposed metrics and the ground truth in the FIRE dataset. a) is DICE, b) is IoU, c) is IoM, d) is SM, e) is SSIM and f) is LPIPS}
    \label{fig:corr}
\end{figure}

The correlation of these metrics with the control point distance metric used  to calculate the Registration Score in FIRE (which has a ground truth) can be found grahically represented in Figure \ref{fig:corr} and analytically measured in Table \ref{tab:corrs}.

\begin{table}[]
\centering
\resizebox{0.35\textwidth}{!}{%
\begin{tabular}{@{}ccc@{}}
\toprule
Metric & \begin{tabular}[c]{@{}c@{}}Spearman \\ correlation\end{tabular} & \begin{tabular}[c]{@{}c@{}}Kendall’s\\  tau\end{tabular} \\ \midrule
DICE & -0.928 & -0.769 \\
IoU & -0.928 & -0.769 \\
IoM & -0.933 & -0.780 \\
SSIM & -0.873 & -0.685 \\
Structure (SM) & -0.878 & -0.692 \\
LPIPS & -0.493 & -0.333 \\ \bottomrule
\end{tabular}%
}
\caption{Correlation metrics between control point distance and the different surrogate metrics proposed}
\label{tab:corrs}
\end{table}

\section{Results and discussion}

\subsection{Loss comparison}
In this section we compare the different variations of our approach (i.e loss functions) in the different datasets: the standard benchmark dataset FIRE \cite{fire} and the new benchmarks LongDRS \cite{ldrs} and DeepDRiD \cite{deepdrid}.

\subsubsection{FIRE}

The results for the different losses in the FIRE dataset \cite{fire} in terms of Registration Score and VTKRS are shown in Tables \ref{tab:fire_l} and \ref{tab:fire_vl}, respectively. Both tables show that AP loss produces the best results in this dataset and, in terms of Registration Score, it does so by a significant margin. This is specially notable in categories A and P. Meanwhile, in category S, the results are more similar across methods, and SP achieves the highest Registration Score by a very small margin over AP. 

Using VTKRS as metric for the results, the performance of the losses is consistent with that of Registration Score. Particularly, AP is still the best loss, while, in this case, SP and NCE tie over the second spot.  Considering the complete dominance of AP in all of the cases, we can say that AP loss produces the best results in the FIRE dataset.


\begin{table}
\centering
\begin{tabular}{@{}lllllll@{}}
\toprule
                           & FIRE           & A              & P              & S              & Avg.           & W. Avg.        \\ \midrule

{Proposed SP}       & {0.755} & {0.760}  & {0.477} & \textbf{{0.946}}         & {0.728} & {0.755} \\

{Proposed NCE}      & {0.758} & {0.749} & {0.489} & {0.945} & {0.728} & {0.758} \\
{Proposed NP}       & {0.576} & {0.546}  & {0.157} & {0.872} & {0.525} & {0.576} \\
{{Proposed AP}} & \textbf{{0.764}} & \textbf{{0.766}} & \textbf{{0.503}} & {0.945} & \textbf{{0.738}} & \textbf{{0.765}}      
      \\ \bottomrule
\end{tabular}%

\caption{Comparison between the different tested losses in the FIRE dataset, measured in Registration Score, sorted by average. Best results highlighted in bold. }
\label{tab:fire_l}
\end{table}

\begin{table}[]
\centering
\begin{tabular}{@{}lllllll@{}}
\toprule
    & FIRE           & A              & P              & S              & Avg.           & W. Avg.        \\ \midrule
SP  & 0.656          & 0.685          & 0.368          & 0.849          & 0.634          & 0.656          \\
NCE & 0.656          & 0.692          & 0.362          & 0.852          & 0.635          & 0.656          \\
NP  & 0.483          & 0.431          & 0.107          & 0.753          & 0.43           & 0.483          \\
AP  & \textbf{0.665} & \textbf{0.696} & \textbf{0.382} & \textbf{0.854} & \textbf{0.644} & \textbf{0.665} \\ \bottomrule
\end{tabular}%
\caption{Comparison of our methods using VTKRS as metric. Best results highlighted in bold.}
\label{tab:fire_vl}
\end{table}

\subsubsection{DeepDRiD}
\label{ssec:ddrid}

The results for the different losses in the DeepDRiD dataset \cite{deepdrid} are shown in Tables \ref{tab:dridl} and \ref{tab:drid_2l}.

Table \ref{tab:dridl} shows the raw results,, meaning that the metrics are influenced by the number of registered pairs. Meanwhile, Table \ref{tab:drid_2l} shows the normalized results, taking into account the number of registrations.


In the DeepDRiD dataset, the best results correspond to the AP loss. This loss registers the highest number of pairs and 
obtains the best metrics in both the un-normalized metrics and the normalized ones. This is even clearer in the normalized Table \ref{tab:drid_2l}. Due to the highest number of registered pairs, the AP loss performs the best in each and every metric.

\begin{table}[]
\centering

\begin{tabular}{lccccccc}
\hline
    & \multicolumn{1}{l}{\# Pairs} & IoU   & DICE  & IoM   & SM    & SSIM  & LPIPS \\ \hline
SP  & 899                          & 0.521 & 0.337 & 0.715 & 0.659 & 0.642 & \textbf{0.183} \\
NCE & 902                          & \textbf{0.526} & 0.339 & \textbf{0.72}  & 0.659 & \textbf{0.643} & \textbf{0.183} \\
NP  & 641                          & 0.442 & 0.295 & 0.621 & 0.646 & 0.634 & 0.194 \\
AP  & \textbf{905}                         & \textbf{0.526} & \textbf{0.34}  & \textbf{0.72}  & \textbf{0.66}  & \textbf{0.643} & \textbf{0.183} \\ \hline
\end{tabular}%

\caption{Results for the different tested loss functions in the DeepDRiD dataset}
\label{tab:dridl}
\end{table}

\begin{table}[]
\centering

\begin{tabular}{lccccccc}
\hline
    & \# Pairs & IoU   & DICE  & IoM   & SM    & SSIM  & LPIPS \\ \hline
SP  & 899      & 0.473 & 0.306 & 0.649 & 0.598 & 0.583 & 0.258 \\
NCE & 902      & 0.479 & 0.309 & 0.656 & 0.600 & 0.586 & 0.256 \\
NP  & 641      & 0.286 & 0.191 & 0.402 & 0.418 & 0.41  & 0.478 \\
AP  & \textbf{905}      & \textbf{0.481} & \textbf{0.311} & \textbf{0.658} & \textbf{0.603} & \textbf{0.588} & \textbf{0.253} \\
\bottomrule
\end{tabular}%

\caption{Normalized results for the different tested loss functions in the DeepDRiD dataset, using the amount of registered pairs.}
\label{tab:drid_2l}
\end{table}

\begin{figure}
    \centering

        \begin{subfigure}{0.47\textwidth}
            \centering
            \includegraphics[width=\textwidth]{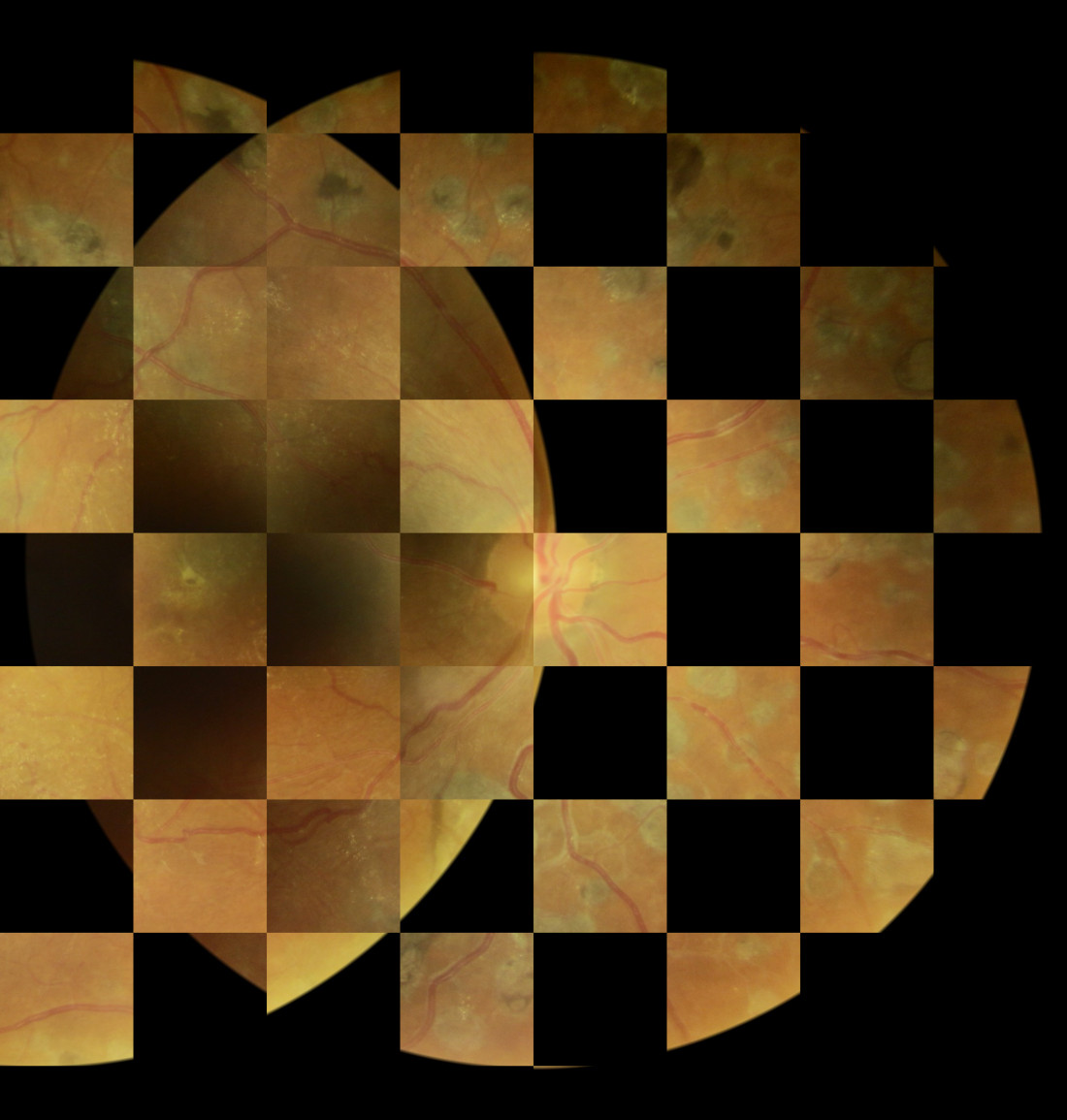}
            \caption{}
            \label{fig:drid_a}
        \end{subfigure}
        \begin{subfigure}{0.47\textwidth}
            \centering
            \includegraphics[width=\textwidth]{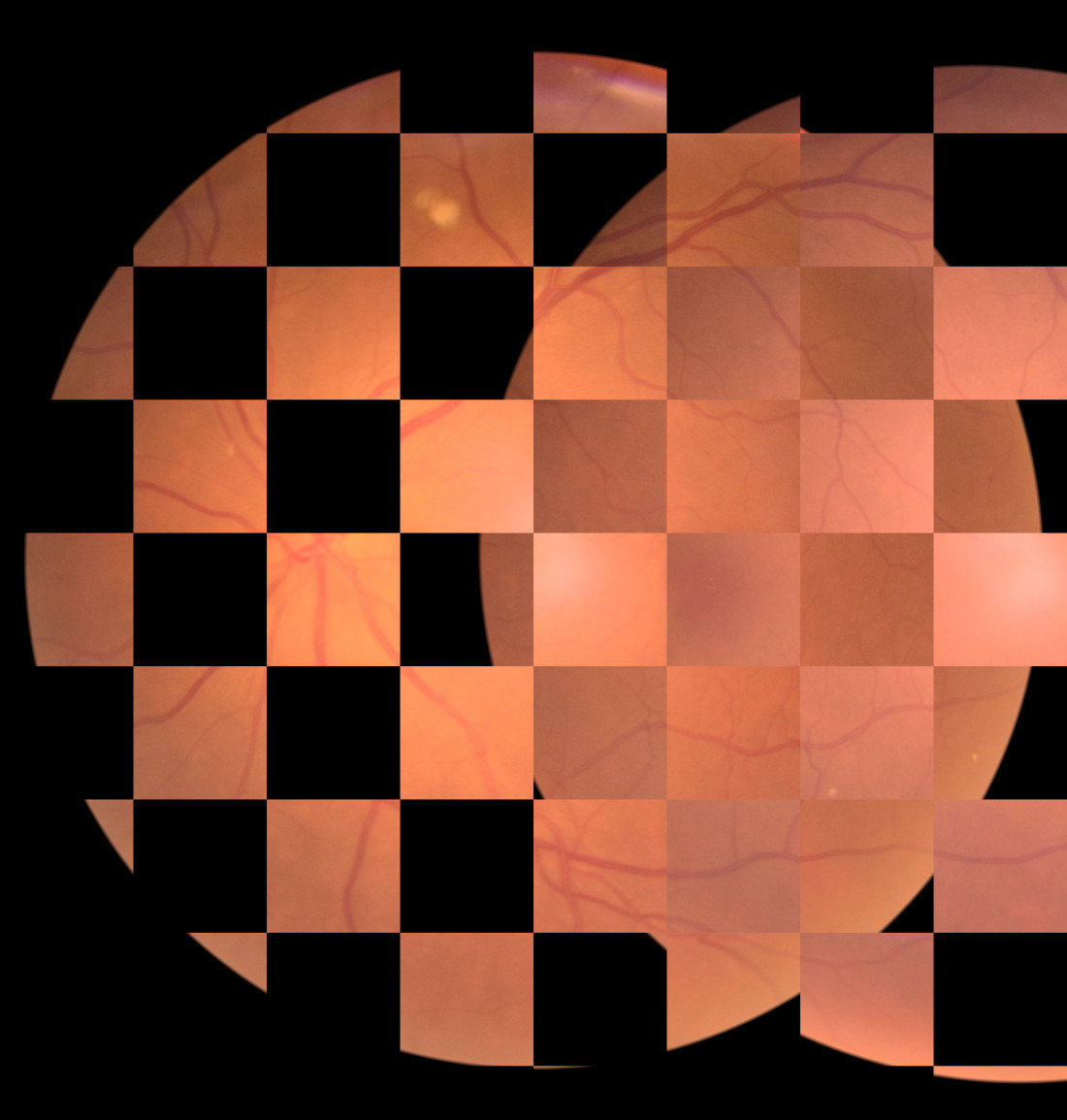}
            \caption{}
            \label{fig:drid_b}
        \end{subfigure}
        \\
        \begin{subfigure}{0.47\textwidth}
            \centering
            \includegraphics[width=\textwidth]{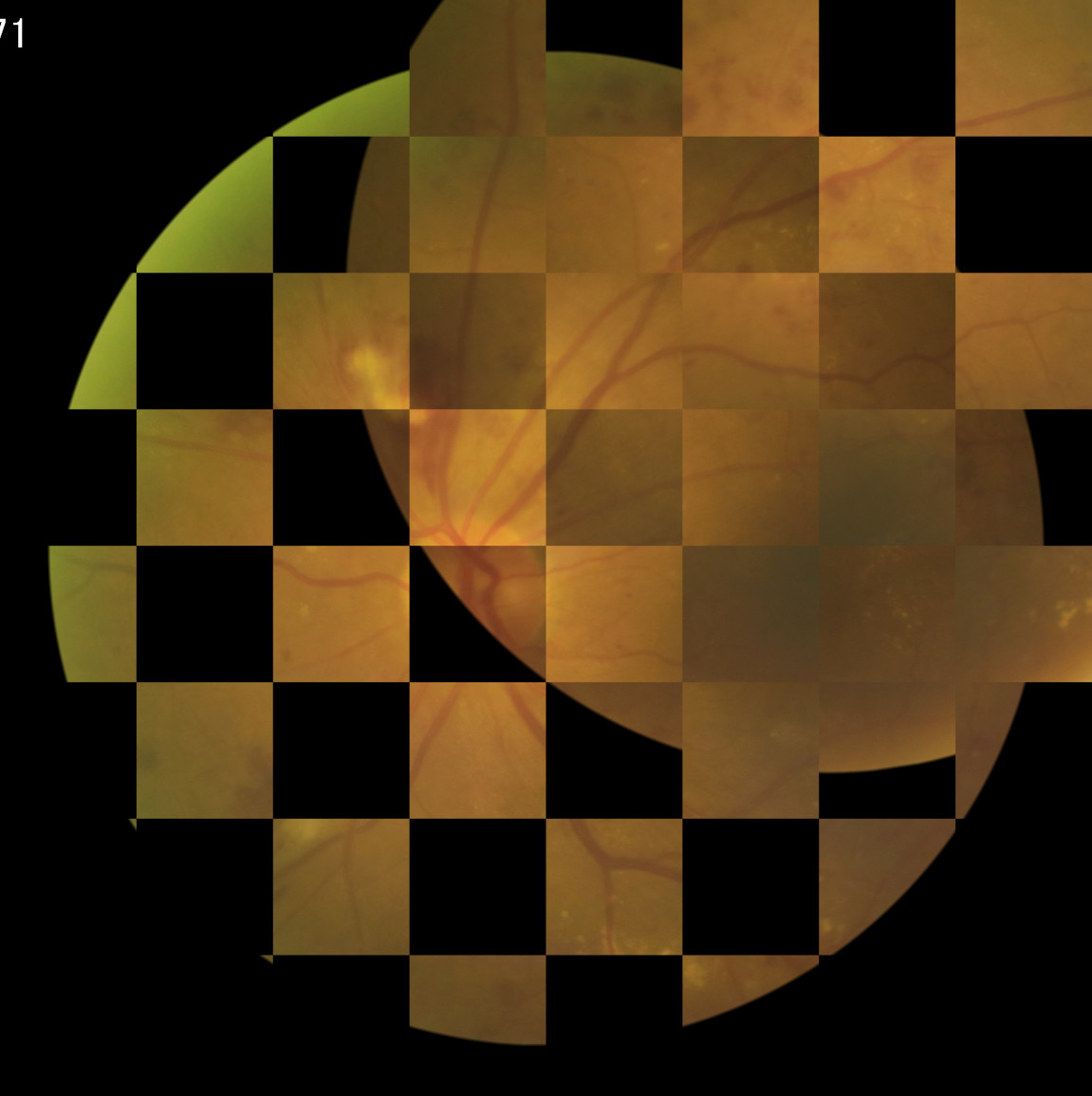}
            \caption{}
            \label{fig:drid_c}
        \end{subfigure}
            \begin{subfigure}{0.47\textwidth}
            \centering
            \includegraphics[width=\textwidth]{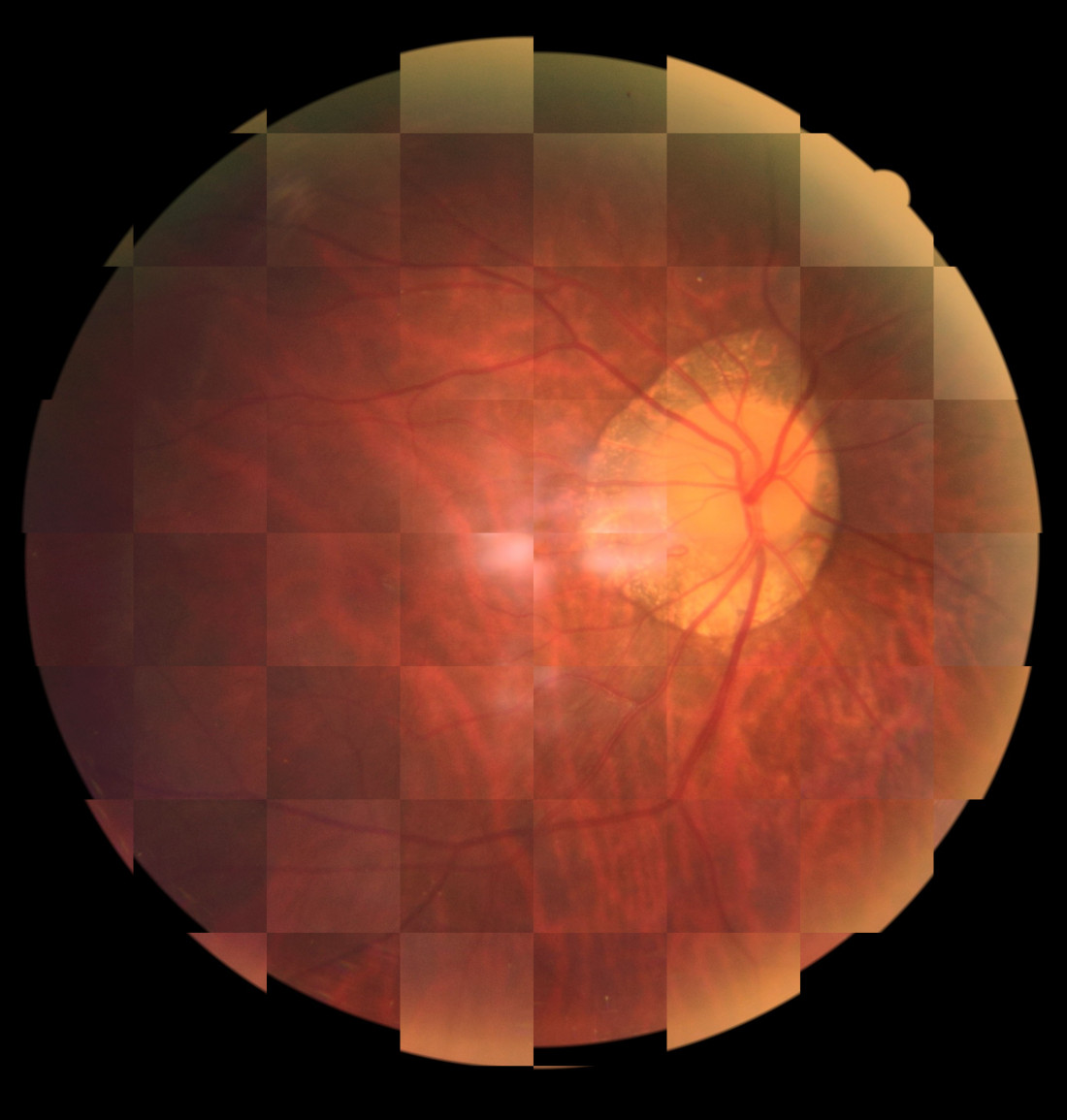}
            \caption{}
            \label{fig:drid_d}
        \end{subfigure}

    \caption{Representative example images from the DeepDRiD dataset, registered using our method trained with AP loss. Each image corresponds to a registration pair from a different patient.}
    \label{fig:res_drid}
\end{figure}

Finally, Figure \ref{fig:res_drid} shows representative images from the DeepDRiD dataset registered using our method with the AP loss.

\subsubsection{LongDRS}

The results for the different losses in the LongDRS dataset \cite{ldrs} are shown in Table \ref{tab:ldrsl}. As previously mentioned in Section \ref{ssec:ddrid} (DeepDrid results), these metrics are un-normalized, therefore we also compute the normalized metrics, taking into account the number of registered pairs, in Table \ref{tab:ldrs_2l}.

Similarly to the DeepDrid dataset, the AP loss achieves the best overall results but ties with other losses in the un-normalized table. However, as seen with the DeepDRiD dataset, the  differences become clearer in the normalized table. In this case, the advantage of AP over NCE is even greater due to the higher difference in number of registered pairs (a difference of 59 pairs in LongDRS compared to just 3 in DeepDrid). Therefore, AP unequivocally demonstrates the best performance.

\begin{table}[]
\centering
\begin{tabular}{lccccccc}
\hline
    & \# Pairs & IoU   & DICE  & IoM   & SM    & SSIM  & LPIPS \\ \hline
SP  & 2688     & 0.572 & 0.357 & 0.742 & \textbf{0.698} & \textbf{0.691} & 0.106 \\
NCE & 2762     & 0.575 & 0.359 & 0.745 & \textbf{0.698} & 0.690 & 0.103 \\
NP  & 1393     & 0.523 & 0.325 & 0.674 & 0.691 & 0.686 & 0.166 \\
AP  & \textbf{2821}     & \textbf{0.577} & \textbf{0.360} & \textbf{0.749} & \textbf{0.698} & 0.690 & \textbf{0.102} \\
\hline

\end{tabular}%
\caption{Results for the different tested loss functions in the LongDRS dataset, both intra-visit and inter-visit. In all the metrics except LPIPS highter is better. Best results highlighted in bold.}
\label{tab:ldrsl}
\end{table}

\begin{table}[]
\centering
\begin{tabular}{lccccccc}
\hline
    & \# Pairs & IoU   & DICE  & IoM   & SM    & SSIM  & LPIPS \\ \hline
SP  & 2688     & 0.49  & 0.306 & 0.635 & 0.597 & 0.591 & 0.235 \\
NCE & 2762     & 0.506 & 0.316 & 0.655 & 0.614 & 0.607 & 0.211 \\
NP  & 1393     & 0.232 & 0.144 & 0.299 & 0.306 & 0.304 & 0.63  \\
AP  & \textbf{2821}     & \textbf{0.518} & \textbf{0.323} & \textbf{0.673} & \textbf{0.627} & \textbf{0.62}  & \textbf{0.193} \\
\hline
\end{tabular}%
\caption{Normalized results for the different tested loss functions in the LongDRS dataset, both intra-visit and inter-visit. In all the metrics except LPIPS highter is better. Best results highlighted in bold.}
\label{tab:ldrs_2l}
\end{table}

LongDRS allows for deeper evaluations of the performance as we can check the registrations in a intra-visit and inter-visit manner. Firstly, the results for the inter-visit registrations are shown in Tables \ref{tab:ldrs_interl} and \ref{tab:ldrs_inter_2l}, in a un-normalized and normalized way, respectively. As in the global evaluation, AP loss obtains the best performance both in the un-normalized and the normalized metrics, as it registers more pairs and does so better than the rest of the losses.

\begin{table}[]
\centering
\begin{tabular}{lccccccc}
\hline
    & \# Pairs & IoU   & DICE  & IoM   & SM    & SSIM  & LPIPS \\ \hline
SP  & 1595     & 0.597 & 0.366 & 0.761 & \textbf{0.707} & 0.699 & 0.121 \\
NCE & 1635     & 0.598 & 0.368 & 0.764 & 0.706 & 0.698 & 0.118 \\
NP  & 945      & 0.567 & 0.345 & 0.714 & 0.706 & \textbf{0.7}   & 0.172 \\
AP  & \textbf{1663}     & \textbf{0.601} & \textbf{0.369} & \textbf{0.768} & 0.706 & 0.698 & \textbf{0.117} \\
\hline
\end{tabular}%
\caption{Inter-visit results for the different different tested loss functions in the LongDRS dataset. In all the metrics except LPIPS highter is better. Best results highlighted in bold.}
\label{tab:ldrs_interl}
\end{table}

\begin{table}[]
\centering
\begin{tabular}{lccccccc}
\hline
    & \# Pairs & IoU   & DICE  & IoM   & SM    & SSIM  & LPIPS \\ \hline
SP  & 1595     & 0.518 & 0.317 & 0.66  & 0.613 & 0.606 & 0.238 \\
NCE & 1635     & 0.532 & 0.327 & 0.679 & 0.628 & 0.621 & 0.216 \\
NP  & 945      & 0.291 & 0.177 & 0.367 & 0.363 & 0.36  & 0.575 \\
AP  & \textbf{1663}     & \textbf{0.543} & \textbf{0.334} & \textbf{0.694} & \textbf{0.638} & \textbf{0.631} & \textbf{0.202} \\
 \hline
\end{tabular}%
\caption{Normalized inter-visit results for the different different tested loss functions in the LongDRS dataset. In all the metrics except LPIPS highter is better. Best results highlighted in bold.}
\label{tab:ldrs_inter_2l}
\end{table}

The intra-visit results for LongDRS are shown in Tables \ref{tab:ldrs_intral} and \ref{tab:ldrs_intra_2l} which display the raw and the normalized metrics, respectively. In this evaluation, the results align with those of the previous experiments; AP registers more pairs and even achieves better raw metrics, making the difference in the normalized metrics even more notable.

\begin{table}[]
\centering
\begin{tabular}{lccccccc}
\hline
    & \# Pairs & IoU   & DICE  & IoM   & SM    & SSIM  & LPIPS \\ \hline
SP  & 1093     & 0.537 & 0.344 & 0.713 & \textbf{0.685} & \textbf{0.678} & 0.084 \\
NCE & 1127     & 0.540 & 0.346 & 0.718 & \textbf{0.685} & \textbf{0.678} & 0.081 \\
NP  & 448      & 0.431 & 0.283 & 0.589 & 0.661 & 0.656 & 0.153 \\
AP  & \textbf{1159}     & \textbf{0.544} & \textbf{0.348} & \textbf{0.722} & \textbf{0.685} & \textbf{0.678} & \textbf{0.080} \\
\hline
\end{tabular}%
\caption{Intra-visit results for the different different tested loss functions in the LongDRS dataset. In all the metrics except LPIPS highter is better. Best results highlighted in bold.}
\label{tab:ldrs_intral}
\end{table}

\begin{table}[]
\centering
\begin{tabular}{lccccccc}
\hline
    & \# Pairs & IoU   & DICE  & IoM   & SM    & SSIM  & LPIPS \\ \hline
SP  & 1093     & 0.451 & 0.289 & 0.599 & 0.575 & 0.569 & 0.456 \\
NCE & 1127     & 0.467 & 0.299 & 0.621 & 0.593 & 0.587 & 0.437 \\
NP  & 448      & 0.148 & 0.097 & 0.203 & 0.227 & 0.226 & 0.794 \\
AP  & \textbf{1159}     & \textbf{0.484} & \textbf{0.310} & \textbf{0.643} & \textbf{0.610} & \textbf{0.604} & \textbf{0.420} \\
\hline
\end{tabular}%
\caption{Normalized Intra-visit results for the different different tested loss functions in the LongDRS dataset. In all the metrics except LPIPS highter is better. Best results highlighted in bold.}
\label{tab:ldrs_intra_2l}
\end{table}

\begin{figure}
    \centering

        \begin{subfigure}{0.32\textwidth}
            \centering
            \includegraphics[width=\textwidth]{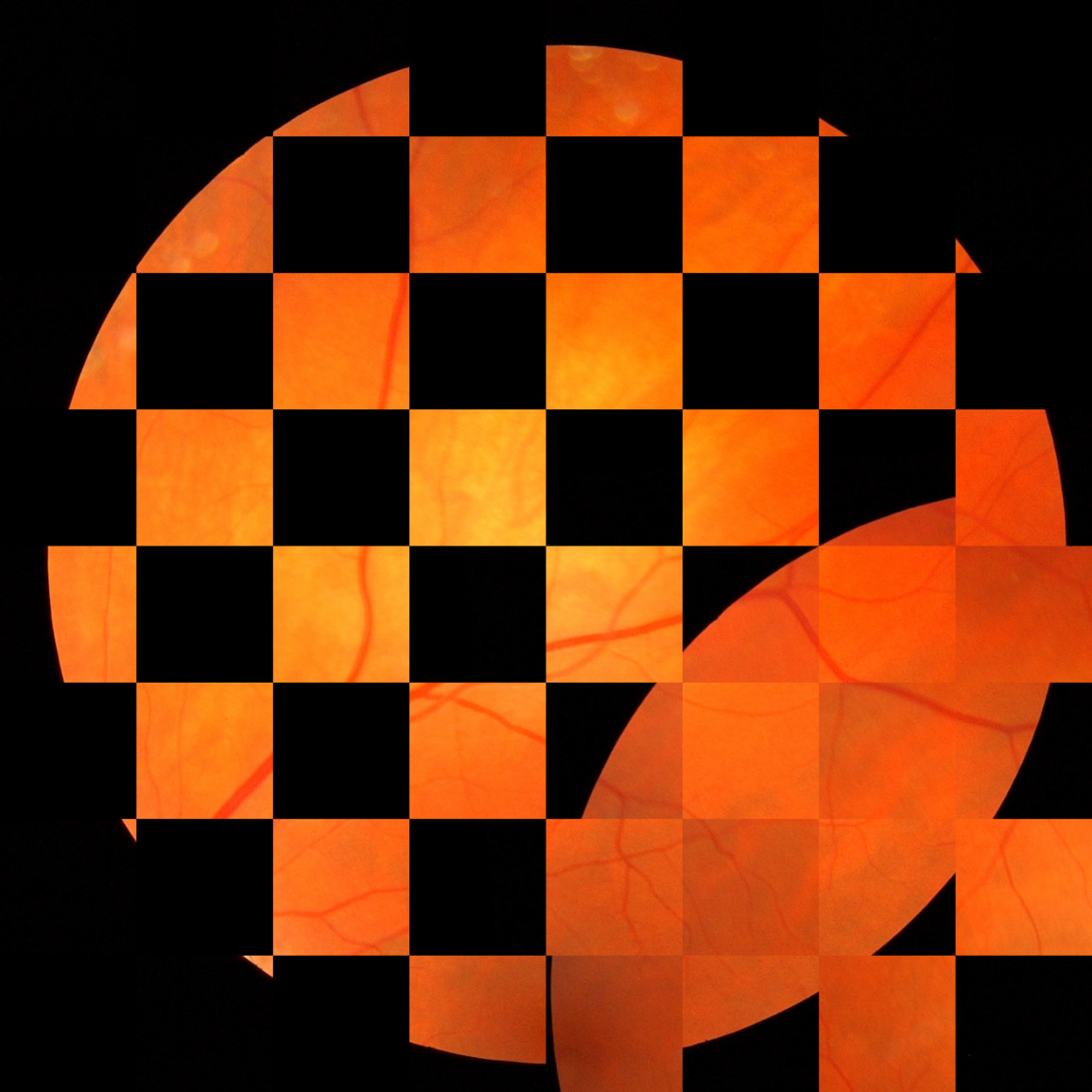}
            \caption{}
            \label{fig:ldrs_a}
        \end{subfigure}
        \begin{subfigure}{0.32\textwidth}
            \centering
            \includegraphics[width=\textwidth]{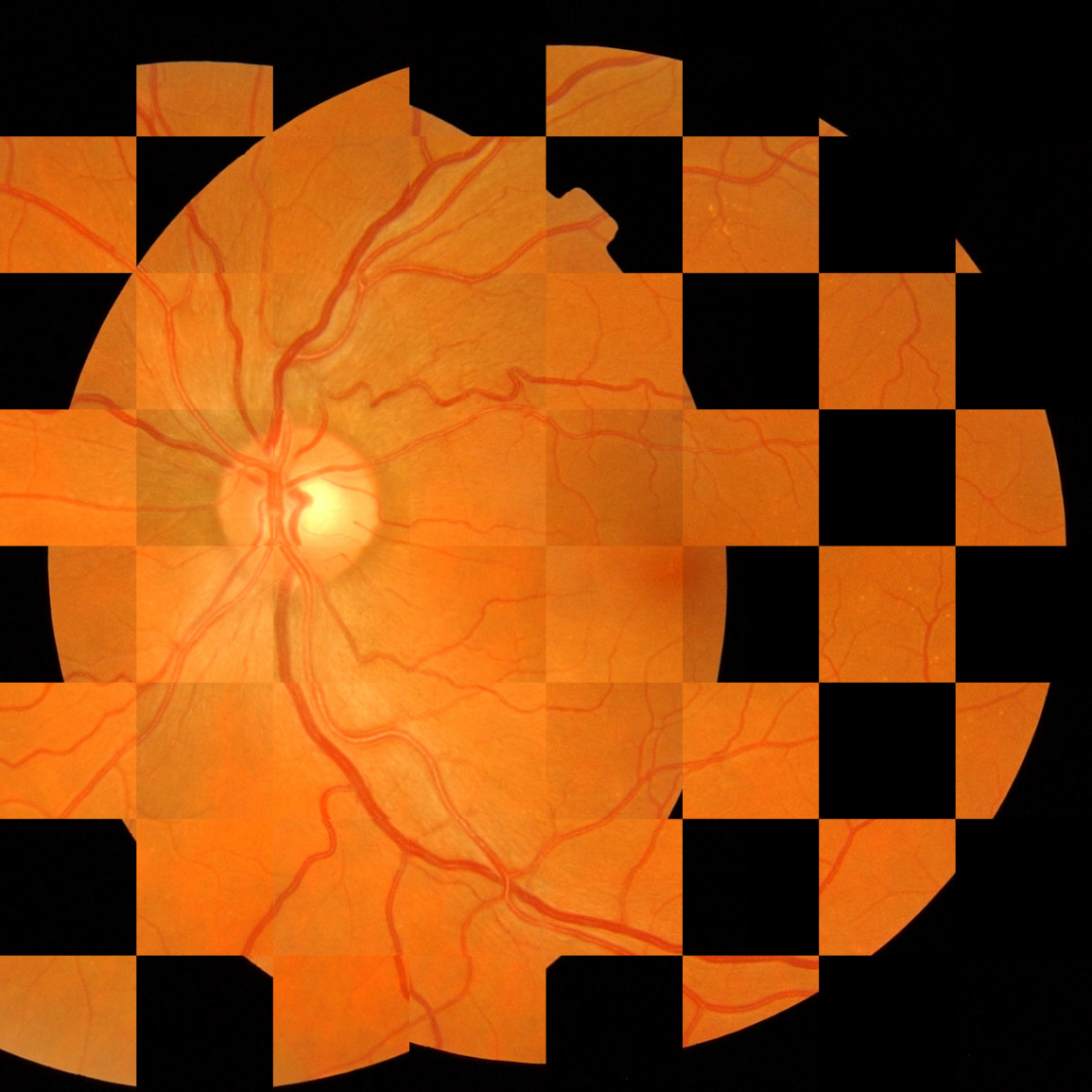}
            \caption{}
            \label{fig:ldrs_b}
        \end{subfigure}
        \begin{subfigure}{0.32\textwidth}
            \centering
            \includegraphics[width=\textwidth]{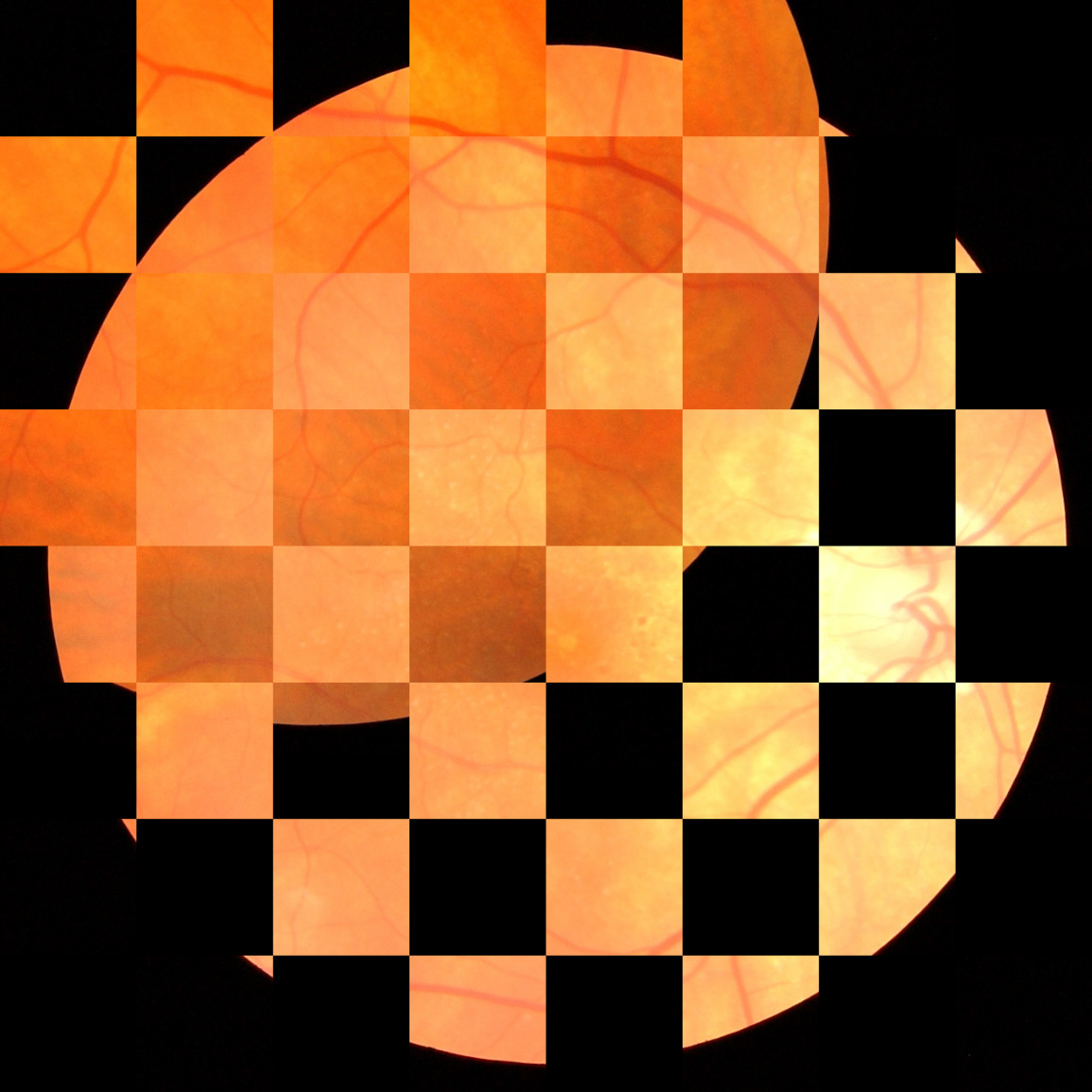}
            \caption{}
            \label{fig:ldrs_c}          
        \end{subfigure}
\\
        \begin{subfigure}{0.32\textwidth}
            \centering
            \includegraphics[width=\textwidth]{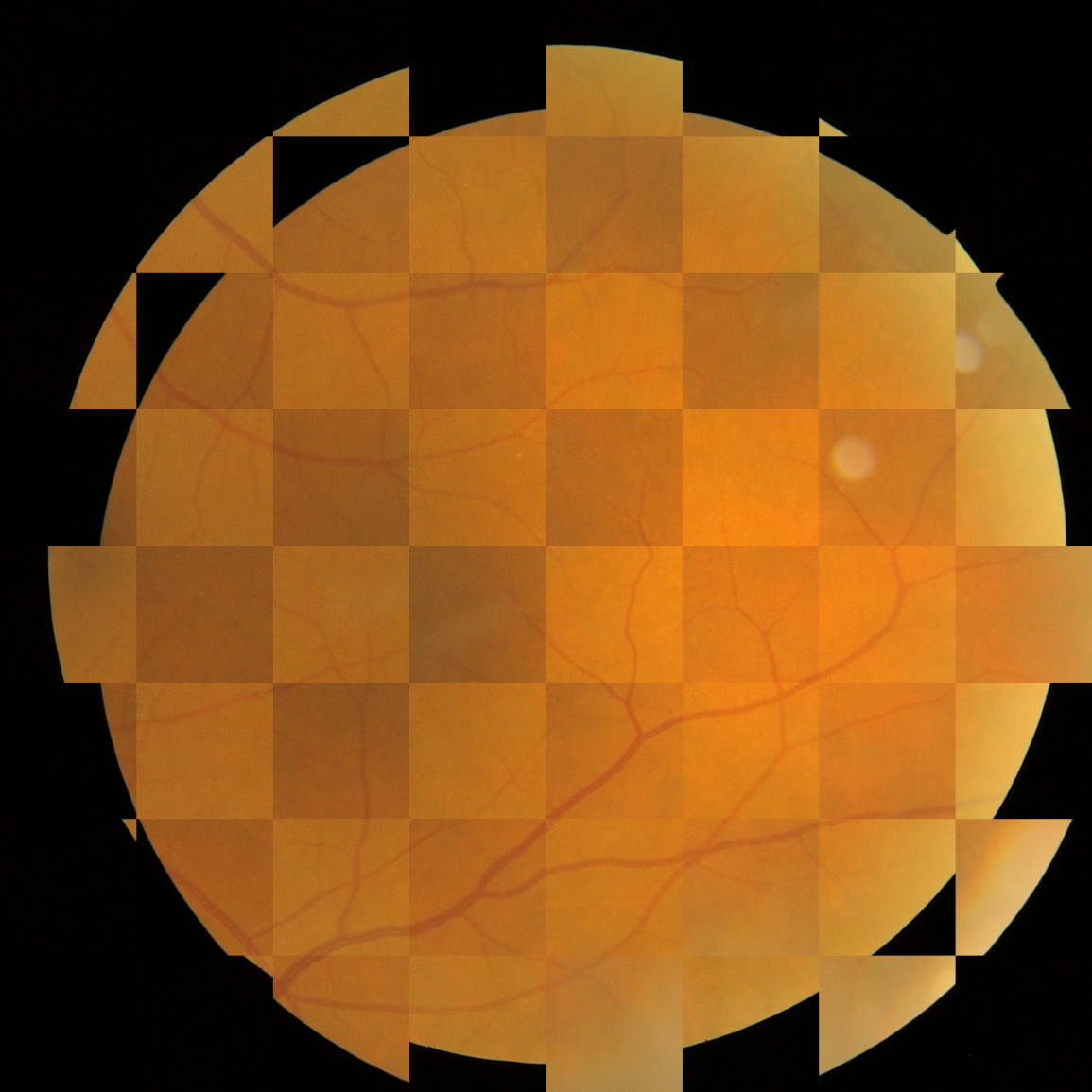}
            \caption{}
            \label{fig:ldrs_d}
        \end{subfigure}
        \begin{subfigure}{0.32\textwidth}
            \centering
            \includegraphics[width=\textwidth]{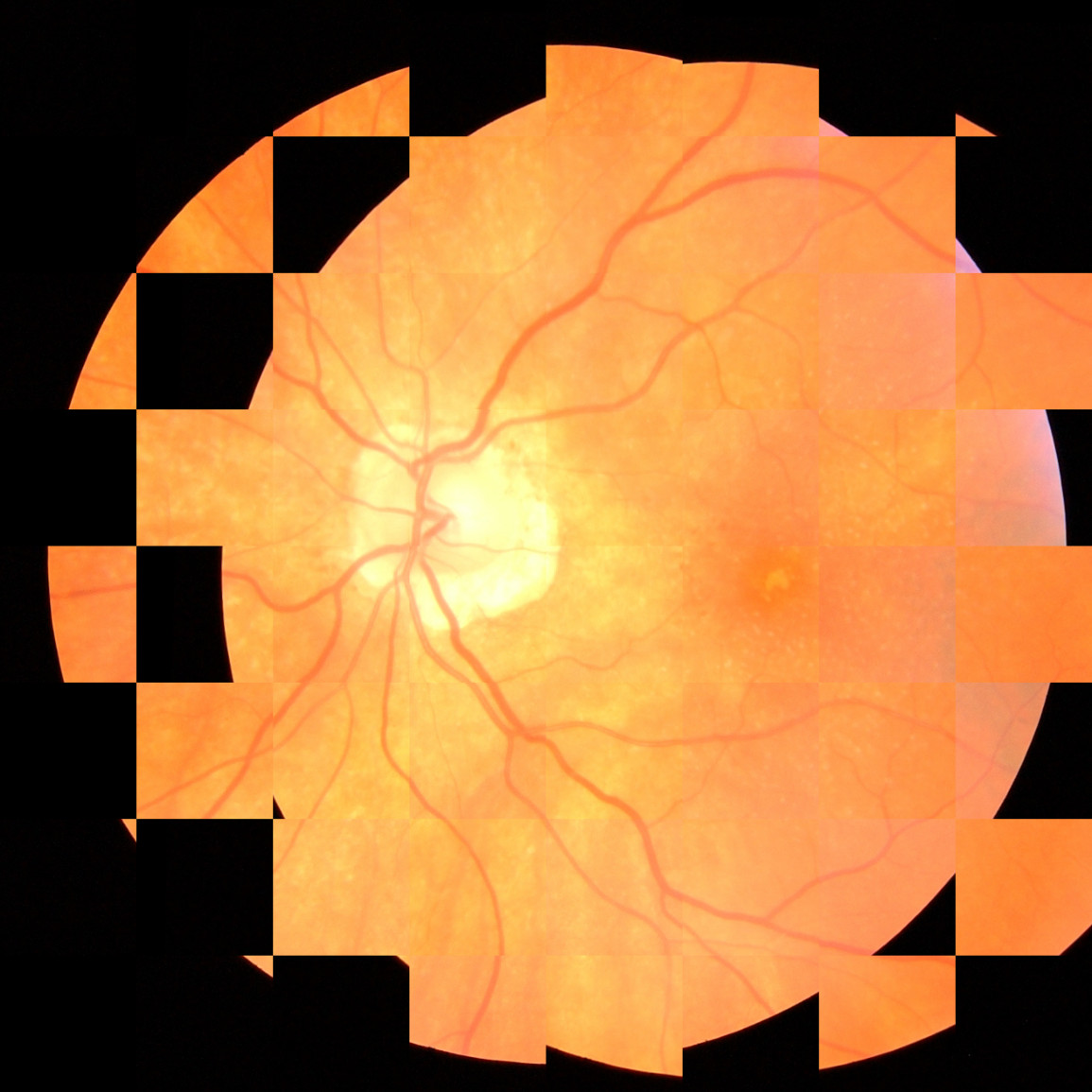}
            \caption{}
            \label{fig:ldrs_e}
        \end{subfigure}
        \begin{subfigure}{0.32\textwidth}
            \centering
            \includegraphics[width=\textwidth]{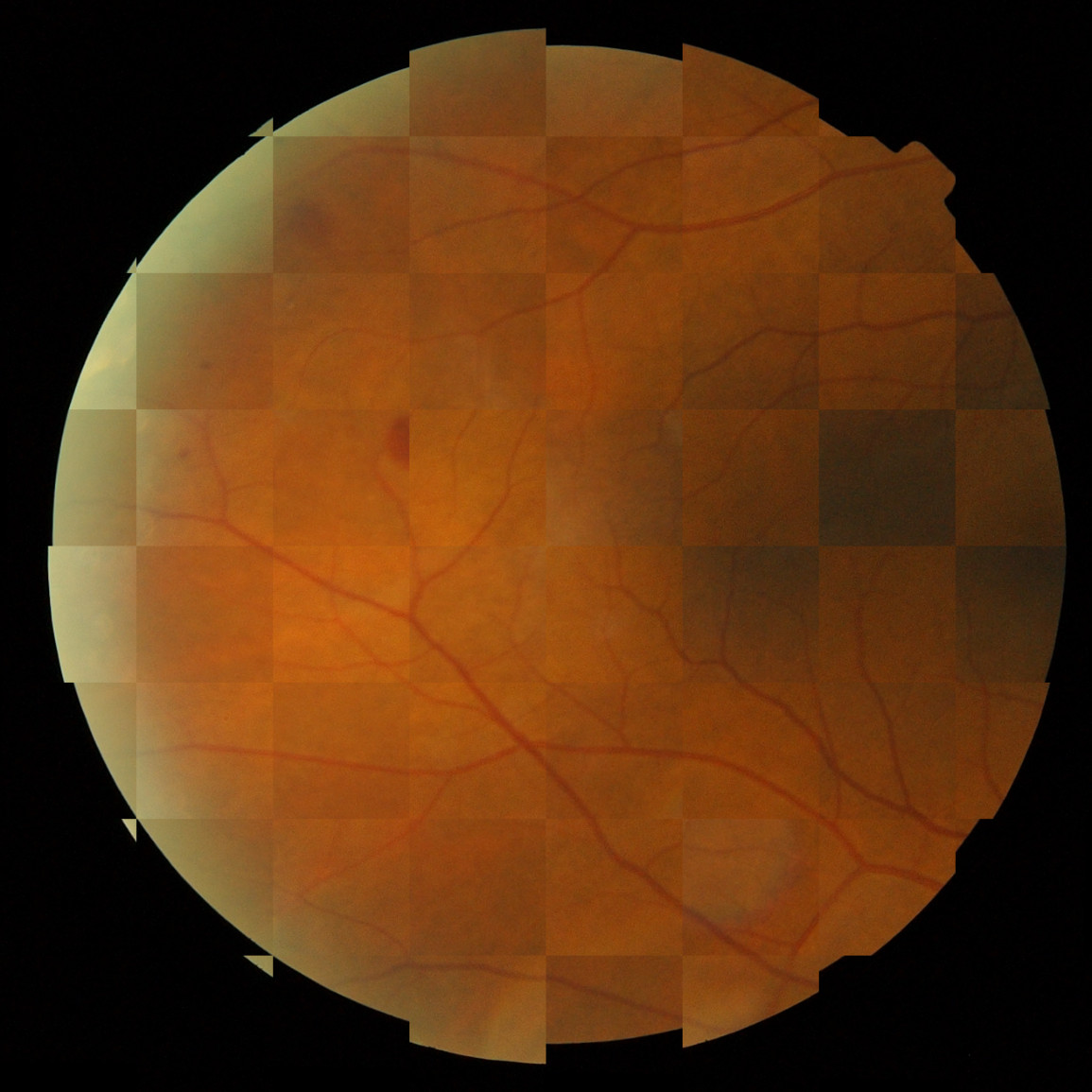}
            \caption{}
            \label{fig:ldrs_f}          
        \end{subfigure}
\\
        \begin{subfigure}{0.32\textwidth}
            \centering
            \includegraphics[width=\textwidth]{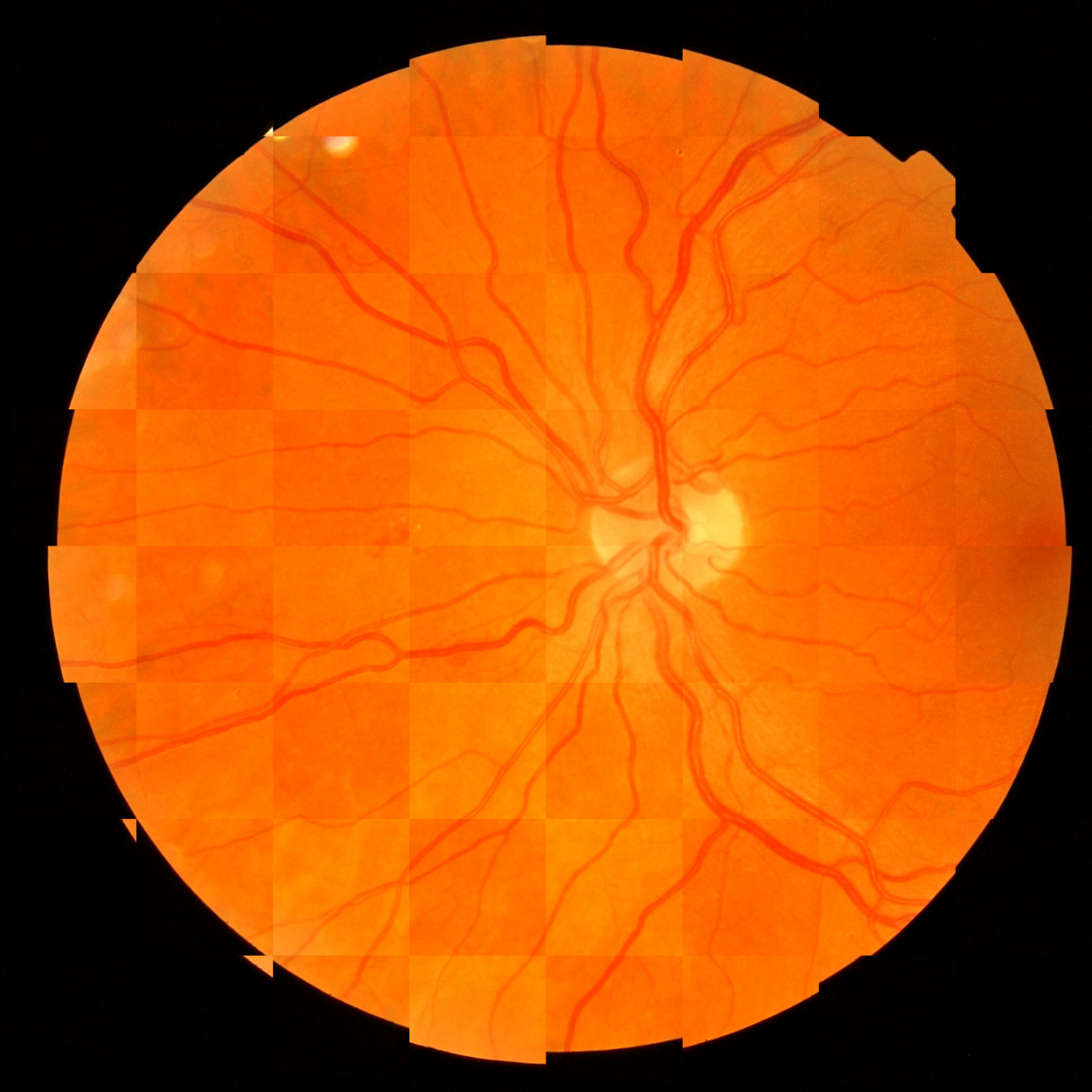}
            \caption{}
            \label{fig:ldrs_g}
        \end{subfigure}
        \begin{subfigure}{0.32\textwidth}
            \centering
            \includegraphics[width=\textwidth]{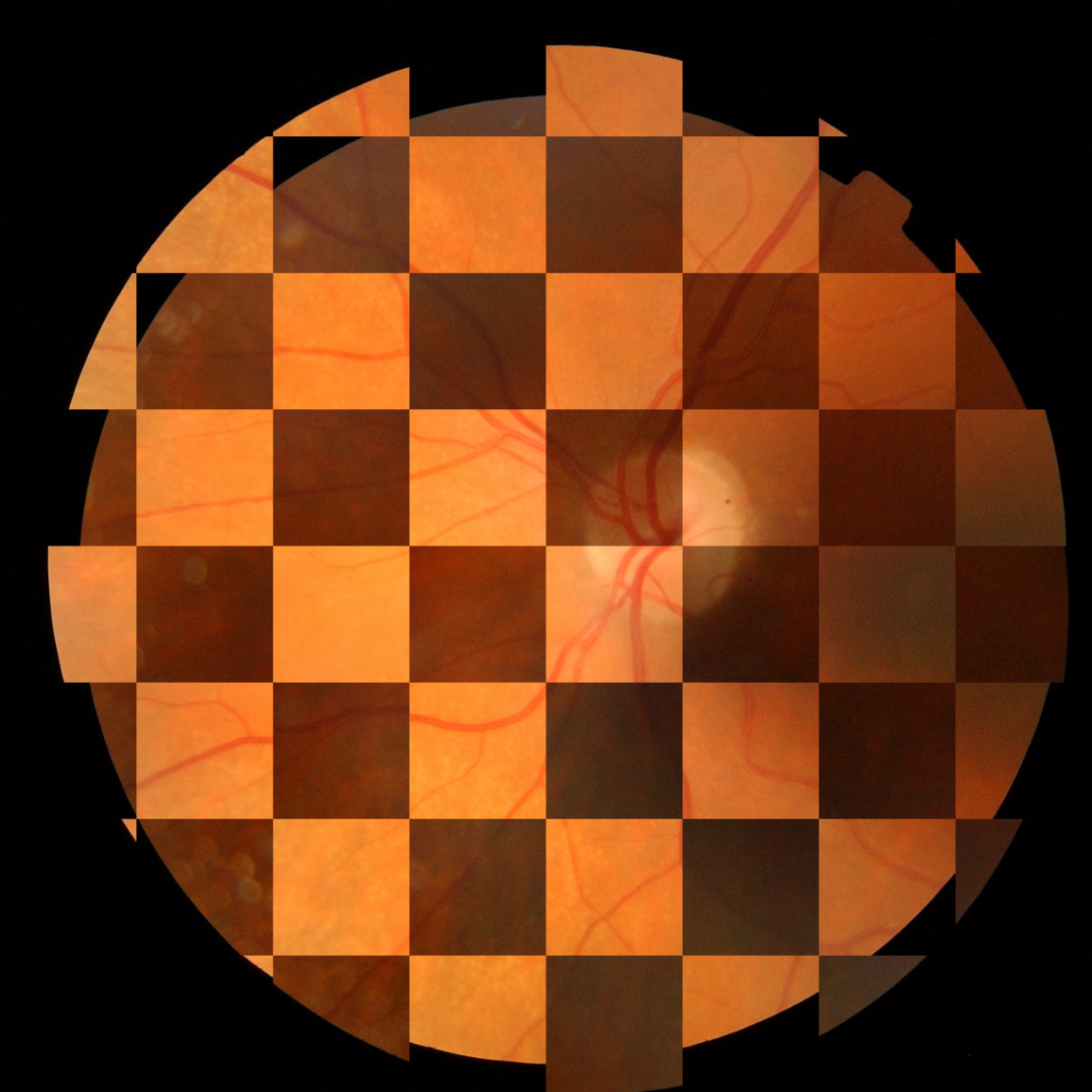}
            \caption{}
            \label{fig:ldrs_h}
        \end{subfigure}
        \begin{subfigure}{0.32\textwidth}
            \centering
            \includegraphics[width=\textwidth]{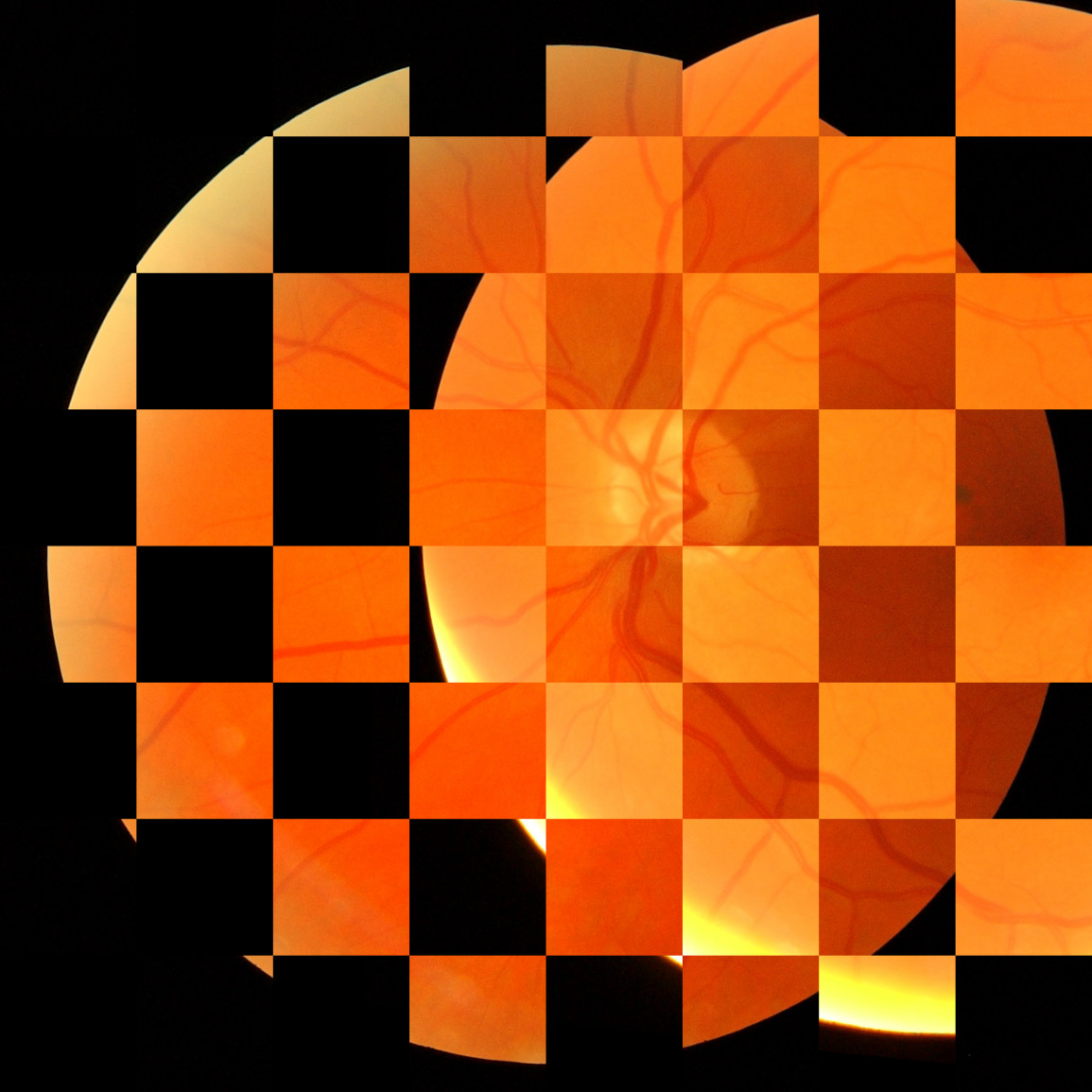}
            \caption{}
            \label{fig:ldrs_i}          
        \end{subfigure}

    \caption{Representative example images from the LongDRS dataset, registered using our method trained with AP loss. a, b and c are intra-visit registration while the rest are inter-visit registrations. Each image corresponds to a registration pair from a different patient.}
    \label{fig:res_ldrs}
\end{figure}

Finally, Figure \ref{fig:res_ldrs} shows representative images from the LongDRS dataset registered using our approach trained using the AP loss.

\subsubsection{Discussion}

Overall, in every dataset, the best results are obtained using the AP loss. This loss has a different behavior than the other three as it creates an implicit ordering and structure in the descriptor space instead of focusing on pulling samples close or apart constantly. This helps the matching process as evidenced by the improved results across all datasets. This is specially notable on DeepDRiD and LongDRS where it registers more pairs and does so more accurately than the other losses.

In terms of the FIRE dataset, AP loss produces the best results across the different evaluations. As for the other three approaches NCE offers the best results closely followed by SupCon loss. N-Pair loss is, by a significant margin, the worst performing loss. 

NCE and SP trade the second place across the FIRE dataset, depending on the particular category. For instance, SP performs better in categories A and S while NCE obtains better results in Category P. However, the difference between the results obtained by these two losses and the results obtained by AP in this category should be highlighted. Due to the lower overlap, an accurate matching process is of utmost importance in category P. Incorrectly matching or not matching a keypoint might significantly worsen the results since there are fewer expected matches. Therefore, category P represents the worse-case scenario within FIRE, specially if, like our method, the number of detections is limited. Thus, the peformance difference between AP and the rest is more marked in this dataset since a mismatched or not-matched keypoint is less likely to be replaced by a nearby one that can substitute it in creaing an accurate transformation. Similarly, category A is expected to have occluded or missing keypoints due to disease progression, however, due to the higher overlapping this is issue mitigated. 

An analysis of pathology progression under lower overlapping (i.e. lower than FIRE category A) can be obtained by using the LongDRS inter-visit evaluation. Using this evaluation as reference we can measure the performance in what would be the worst case scenario in day-to-day clinical practice. In the inter-visit evaluation we can see that AP loss registers significantly more pairs than NCE or SP. In this evaluation, NP is still the worst loss. In this case, the gap between NCE and SP is sufficiently significant to confirm that NCE performs better in this particular evaluation (i.e pathology progression with varied-although-low overlapping), despite performing worse in the Category A of FIRE which also had pathology progression but, overall, bigger overlapping.

As for the rest of the LongDRS (i.e. the intra-visit evaluation), the results are similar. This evaluation allows to quantify the performance of the methods when no pathology progression takes place. Thus, this evaluates only the matching performance using only different viewpoints. This subset has even lower overlapping than the inter-visit evaluation. The results of this evaluation are in line with the inter-visit one as the order in terms of results obtained by the different losses does not vary. AP performs the best, followed by NCE, SP and NP in last place. Finally, for the global evaluation of this dataset, the results are consistent with the two parts of the evaluation already mentioned. AP offers the best performance by a significantly wide margin, highlighted by the 59 extra pairs that it can register over NCE, the second loss in terms of results. Similarly, NCE registers 74 more pairs than SP, solidifying its position as the second loss in terms of results.

Lastly, when using the DeepDRiD dataset, we observe that the differences between the losses are notably smaller compared to those in LongDRS. This discrepancy can be attributed to the larger degree of overlap present in the pairs in DeepDRiD. As the images are intended to be centered on the macula and optic disc, they are expected to exhibit medium or high levels of overlap. However, due to imperfect centering, this is not always the case, resulting in considerable variance in overlapping portions within this dataset. Nonetheless, it generally demonstrates higher levels of overlap compared to LongDRS, albeit lower when compared to FIRE. This higher overlap diminishes the performance disparities among the proposed losses, which was clearer in LongDRS. This is because higher overlapping parts ensure that most keypoints can be matched since they are present in both images of the pair. Consequently, the importance of precise matching is reduced, since even if a particular keypoint is mismatched or not matched at all, there are others available that can still contribute to an accurate registration. Nevertheless, the ranking of the proposed loss functions still remains the same, with AP being the superior approach, NCE ranking second, followed by SP and NP trailing behind by a wide margin.

Overall, the results are consistent from one dataset to the others, evidencing that AP Loss produces the best results. As previously mentioned, the main difference between this approach and the others is the fact that it works in a different way which allows the descriptor space to have an implicit structure and ordering. This proves beneficial as the results show that AP loss can register more pairs (in DeepDRiD and LongDRS) and do so better than the rest of the proposed losses. On the other hand, the three remaining losses make use of the same type of paradigm, pulling closer the positive samples and apart the negative ones. This approach, despite its high popularity, has a significant pitfall. These loss objectives contain no real information about semantical relations between the embeddings. This way, a hard-negative sample with semantic relations to the anchor, will be negatively punished for being too close to it. This, in turn, may be harmful to generate useful descriptors. Instead, AP loss ranks every sample, making sure that the positives ones are first in this rank. Thus, the negative samples also have an ordering between, allowing them to preserve semantic information as the more similar ones should be ranked closer to the anchor.

Despite this, NCE and SP provide accurate performances across all datsets. However, there are significant differences between these two and NP. Overall the NCE loss performs the better than SP and NP. The difference between NCE and SP lies in the inclusion of the positive samples in the denominator. While the denominator in NCE contains only the positive sample from the numerator, in SP the denominator contains effectively all samples, that is all the negative pairs as well as all the positive ones. Thus, for descriptor learning, only including the current anchor-positive sample in the contrast offers better performance. On the other hand, the difference between NCE and NP is the inclusion of the temperature parameter. While in NCE it can be liberally modified in NP it is effectively fixed as 1.

Low temperature parameters increase the penalty on the hardest negative samples (i.e. the most similar to the anchor). As the temperature increases the penalties are more uniform across all available samples. Thus, the lower the temperature is, the more hardness-aware the loss becomes, placing more importance on the negative samples which are the most similar to the anchor (i.e. hard-negatives). Low temperature allows to create uniform embedding distributions which is a desirable property, especially for descriptor matching. However, if the contrastive loss has a small temperature setting, it will create large penalties for the nearest hard-negatives. This will strongly push semantically similar samples apart, breaking the underlying semantic structure of the embedding distribution. Thus, larger temperature parameters allow more tolerance to semantically similar samples at the cost of less relevant contrasts and a less uniform embedding distribution. Thus, optimizing the temperature parameter becomes a tricky process of balancing the uniformity and semantic meaning of the embeddings.

In contrastive approaches with only hard-negative samples the uniformity is kept stable across temperature parameters due to the negative samples themselves. Moreover, hard-negative sampling reduces the effect of the temperature parameter in controlling the hardness-aware property of the loss. Thus, although NP has a large temperature parameter, given that our approach uses only hard-negative samples it could still, theoretically, obtain accurate performance.

The results of NP show that the temperature is key in the performance of this types of losses in spite of the hard-negative-based approach that we use. The lower temperature used in NCE, and SP for that matter, allows to create more uniform embedding spaces which allows for easier discernibility between descriptors facilitating matching. The higher emphasis on the semantic relations resulting from using NP as the loss do not translate to real word registration performance which might be due to the structure of the embedding space that this loss produces. Furthermore, the comparative results of NCE and NP reinforce the notion presented previously in relation to the advantage of AP. The semantic ordering and structure that AP loss provides is key to its performance. Furthermore, as AP does not have a temperature parameter requiring careful finetunning. In this regard, it should be noted that the temperature in NCE nor SP were subject to any tunning at all. We used the values from the reference work \cite{rivas3}. Thus, although it would be desirable to find the better temperature parameter for this particular task and framework, it is out of the scope of this paper since we have already proved the benefits of AP loss. Even if we were to improve NCE or SP it is unlikely to be by a significant enough amount that it would justify adding an extra validation dataset and running the computationally expensive training step multiple times.

Overall, in this paper we tested four losses belonging to two different variations of self-supervised learning. InfoNCE, N-Pair and SupCon Losses are based on the notion of maximizing distances between embeddings. On the other hand, the AP loss just intends to match the anchor with its true positives, striving to obtaining a high average precision. Regarding the performance, as previously stated, the AP loss produces better results than any of the other three. Therefore, at least for descriptor computation, optimizing towards AP rather than just distance is beneficial in terms of results.

\subsection{State of the art comparison}

In this section we compare our best approach, the one that uses AP Loss, against the rest of the state of the art works. 

\subsubsection{FIRE}


\begin{table}[]
\centering
\resizebox{0.7\textwidth}{!}{%
\begin{tabular}{@{}lllllll@{}}
\toprule
                           & FIRE           & A              & P              & S              & Avg.           & W. Avg.        \\ \midrule
VOTUS   \cite{votus}                   & \textbf{0.812} & 0.681          & \textbf{0.672} & 0.934          & \textbf{0.762} & \textbf{0.811} \\
SR Manual  \cite{eccv20}                & -              & 0.783          & 0.542*         & 0.94           & 0.755*         & 0.780*         \\
REMPE \cite{rempe}                  & 0.773          & 0.66           & 0.542          & \textbf{0.958} & 0.72           & 0.774          \\
Retina-R2D2 \cite{rivas2}             & 0.695          & 0.726          & 0.352          & 0.925          & 0.645          & 0.695          \\
Rivas-Villar  \cite{rivas}             & 0.657          & 0.660          & 0.293          & 0.908          & 0.620          & 0.657          \\
ConKeD AP (Ours) & 0.764 & 0.766 & 0.503 & 0.945 & 0.738 & 0.765 \\
\bottomrule
\end{tabular}%
}
\caption{Comparison between the different versions of our approach and state of the art methods, measured in Registration Score, sorted by average. * indicates that the results were calculated using one image less. Best overall results in bold.}
\label{tab:fire}
\end{table}


The results for this experiment are shown in Table \ref{tab:fire}. Overall, VOTUS \cite{votus} is the method that offers the best performance. However, it does not perform in-line with the best methods in Category A, the category with disease progression and thus the most relevant for clinical practice. Our method obtains comparable results to current state of the art methods. In particular, our method is only behind a single method in two categories (S \& A).


\subsubsection{DeepDRiD}

While FIRE is the de-facto benchmark dataset, DeepDRiD and LongDRS have never been used for registration despite having suitable pairs. Thus, no other work has been evaluated in this dataset. However, validating registration methods in multiple datasets is fundamental to confirm that it can work with images from other devices, achieving generalized CF registration. This way, we can only compare our method to works in the state of the art that are public since they do not offer results in these datasets. This leaves only SuperRetina \cite{eccv20} as it is fully open source. The rest of the state of the art methods are either not available or are compiled which does not allow for the necessary modifications. It should be noted that these modifications are just related to how and where the data is read and not related to the method at all. Therefore, it must be highlighted that SuperRetina is not modified in any way for these experiments, it is simply used as provided.

The results for our best approach as well as SuperRetina \cite{eccv20} are shown in Tables \ref{tab:drid} and \ref{tab:drid_2}. The Table \ref{tab:drid} shows the direct, un-processed results, that is, the metrics obtained from the images that each method registers. This allows for comparison of just the metrics in the accurately registered images. On the other hand Table \ref{tab:drid_2} contains the metrics normalized following the number of reasonably transformed images. This Table allows easy comparison among methods as it takes into account the number of accurately registered images.

\begin{table}[]
\centering

\begin{tabular}{lccccccc}
\hline
    & \multicolumn{1}{l}{\# Pairs} & IoU   & DICE  & IoM   & SM    & SSIM  & LPIPS \\ \hline

ConKeD AP  & 905                          & 0.526 & 0.34  & 0.72  & 0.66  & 0.643 & 0.183 \\
SuperRetina  & 979                          & 0.546 & 0.349 & 0.748 & 0.67  & 0.652 & 0.186 \\ \hline
\end{tabular}%

\caption{Results in the DeepDRiD dataset. In all the metrics except LPIPS higher is better.}
\label{tab:drid}
\end{table}

\begin{table}[]
\centering

\begin{tabular}{lccccccc}
\hline
    & \# Pairs & IoU   & DICE  & IoM   & SM    & SSIM  & LPIPS \\ \hline
ConKeD AP  & 905      & 0.481 & 0.311 & 0.658 & 0.603 & 0.588 & 0.253 \\
SuperRetina  & 979      & 0.540 & 0.345 & 0.74  & 0.663 & 0.645 & 0.195 \\ \hline
\end{tabular}%

\caption{Normalized results in the DeepDRiD dataset, using the amount of registered pairs. In all the metrics except LPIPS highter is better.}
\label{tab:drid_2}
\end{table}

\subsubsection{LongDRS}

The results for our method and SuperRetina \cite{eccv20} are shown in Tables \ref{tab:ldrs} and \ref{tab:ldrs_2}, containing the unprocessed and the normalized results for this dataset, respectively. 

Despite the notable differences between the the LongDRS and the DeepDRiD datasets the results are similar. In this case, SuperRetina is able to register a higher number of images and produce the best metrics due to its higher amount of keypoints. Our method is penalized in the normalized Table, due to the lower number of registered pairs, although it maintains a relatively similar performance to SuperRetina.

\begin{table}[]
\centering
\begin{tabular}{lccccccc}
\hline
    & \# Pairs & IoU   & DICE  & IoM   & SM    & SSIM  & LPIPS \\ \hline
ConKeD AP  & 2821     & 0.577 & 0.360 & 0.749 & 0.698 & 0.690 & 0.102 \\
SuperRetina  & 3063     & 0.604 & 0.373 & 0.780 & 0.710 & 0.701 & 0.099 \\ \hline
\end{tabular}%
\caption{Results for our approach and SupeRetina in the LongDRS dataset. In all the metrics except LPIPS highter is better. Best results highlighted in bold.}
\label{tab:ldrs}
\end{table}

\begin{table}[]
\centering
\begin{tabular}{lccccccc}
\hline
    & \# Pairs & IoU   & DICE  & IoM   & SM    & SSIM  & LPIPS \\ \hline
ConKeD AP  & 2821     & 0.518 & 0.323 & 0.673 & 0.627 & 0.62  & 0.193 \\
SuperRetina  & 3063     & 0.589 & 0.364 & 0.761 & 0.692 & 0.684 & 0.121 \\ \hline
\end{tabular}%
\caption{Normalized results our approach and SupeRetina in the LongDRS dataset. In all the metrics except LPIPS highter is better. Best results highlighted in bold.}
\label{tab:ldrs_2}
\end{table}

Due to the structure of this dataset, the results from LongDRS can be divided into intra-visit and inter-visit, as previously explained. The results for inter-visit split of the LongDRS dataset are shown in Tables \ref{tab:ldrs_inter} and \ref{tab:ldrs_inter_2}, containing the raw and normalized results for this dataset, respectively. On the other hand, the results for intra-visit evaluation are shown in Tables \ref{tab:ldrs_intra} and \ref{tab:ldrs_intra_2}.

\begin{table}[]
\centering
\begin{tabular}{lccccccc}
\hline
    & \# Pairs & IoU   & DICE  & IoM   & SM    & SSIM  & LPIPS \\ \hline
ConKeD AP  & 1663     & 0.601 & 0.369 & 0.768 & 0.706 & 0.698 & 0.117 \\
SuperRetina  & 1800     & 0.623 & 0.38  & 0.794 & 0.718 & 0.708 & 0.114 \\ \hline
\end{tabular}%
\caption{Inter Results for our approach and SupeRetina in the LongDRS dataset. In all the metrics except LPIPS highter is better. Best results highlighted in bold.}
\label{tab:ldrs_inter}
\end{table}

\begin{table}[]
\centering
\begin{tabular}{lccccccc}
\hline
    & \# Pairs & IoU   & DICE  & IoM   & SM    & SSIM  & LPIPS \\ \hline
ConKeD AP  & 1663     & 0.543 & 0.334 & 0.694 & 0.638 & 0.631 & 0.202 \\
SuperRetina  & 1800     & 0.61  & 0.372 & 0.777 & 0.703 & 0.693 & 0.133 \\ \hline
\end{tabular}%
\caption{Normalized inter-visit results for  our approach and SupeRetina in the LongDRS dataset. In all the metrics except LPIPS highter is better. Best results highlighted in bold.}
\label{tab:ldrs_inter_2}
\end{table}

\begin{table}[]
\centering
\begin{tabular}{lccccccc}
\hline
    & \# Pairs & IoU   & DICE  & IoM   & SM    & SSIM  & LPIPS \\ \hline
ConKeD AP  & 1159     & 0.544 & 0.348 & 0.722 & 0.685 & 0.678 & 0.080 \\
SuperRetina  & 1263     & 0.578 & 0.364 & 0.759 & 0.699 & 0.69  & 0.078 \\ \hline
\end{tabular}%
\caption{Intra-visit registration results for our approach and SupeRetina in the LongDRS dataset. In all the metrics except LPIPS highter is better.}
\label{tab:ldrs_intra}
\end{table}

\begin{table}[]
\centering
\begin{tabular}{lccccccc}
\hline
    & \# Pairs & IoU   & DICE  & IoM   & SM    & SSIM  & LPIPS \\ \hline
ConKeD AP  & 1159     & 0.484 & 0.310 & 0.643 & 0.610 & 0.604 & 0.420 \\
SuperRetina  & 1263     & 0.561 & 0.353 & 0.736 & 0.678 & 0.669 & 0.367 \\ \hline
\end{tabular}%
\caption{Normalized intra-visit registration results our approach and SupeRetina in the LongDRS dataset. In all the metrics except LPIPS highter is better.}
\label{tab:ldrs_intra_2}
\end{table}

\subsubsection{Discussion}

Overall our method is able to produce accurate registrations, obtaining similar metrics to the best state-of-the-art deep learning method, SuperRetina.


In FIRE, VOTUS \cite{votus} obtains the best overall performance. This method is a classical approach that uses a higher-degree transformation which enables it obtain results in category P that are much higher than any competing method. This comes at the cost of worse performance in categories A and S, which have smaller expected displacements. Similarly, REMPE \cite{rempe} (also a classical method) uses an ad-hoc transformation, made specifically for the eye fundus. However, in category P, which could benefit the most from this domain-specific transformation due to the expected big transformations, it is not able to improve the results of VOTUS.  On the other hand SuperRetina is a deep learning method similar to our own approach. In this regard it uses an homography transformation as well, so the results should directly comparable in this regard. SuperRetina obtains very similar results to our approach although slightly better in all the categories except in S where our method barely improves it. However, the disadvantages of SuperRetina are  that it uses more training samples, requires more keypoints (nearly 5$\times$), requires ad-hoc image pre-processing and double inference step which involves detecting and describing keypoints, matching them, transforming the moving image and the repeating the whole process with the transformed moving image. On the other hand, our method does not pre-process the images and a second inference step does not improve the results, which evidences that our keypoints are more robust and that the descriptors are more invariant to transformations.

Overall, we can conclude that our method provides satisfactory results as it provides results that are competitive with the best state of the art works while having several significant advantages like the use of deep learning, the low keypoint requirement, no pre-processing or double inference, etc.


DeepDRiD and LongDRS allows to validate our approach and confirm its performance on other datasets different than the typical FIRE. Moreover, we also evaluate SuperRetina to obtain a metric of the state of the art.

In DeepDRiD, considering the number of registered image pairs, SuperRetina obtains the best result, registering 979 pairs of the total 990. Our approach is able to register 905 image pairs. Overall, the metrics as well as the successfully registered pairs are specially influenced by the number of detected keypoints. DeepDRiD is mainly composed of registration pairs with varied, although fairly low, overlapping as it is composed of images centered on the optic disc and the macula. However, as it can be seen in Figure \ref{fig:drid_d} there are also pairs with high overlaps. Thus, 
the main complexities of this dataset are its low overlapping as well as the artifacts present in the images. This low overlapping heavily penalizes our method, in comparison with SuperRetina.  Our approach only uses the blood vessel crossovers and bifurcations, therefore the quality of the registration is entirely dependent on how many of these keypoints are visible, at the same time, in the overlapping part of both images. On the other hand, SuperRetina is able to detect more keypoints which makes it more resistant to big deformations by sheer quantity of keypoints. This, as demonstrated by the results in Tables \ref{tab:drid} and \ref{tab:drid_2} makes their approach produce better results. However, it should be noted that the results of our proposed method are remarkably close. Despite the challenging setting, our method comes really close to SuperRetina. Moreover, our approach is limited by the natural amount of crossovers and bifurcations. If the overlapping part of the image has low vascularity or simply a small amount of crossovers and bifurcations our method will inevitably fail. Despite this, while we can conclude that the number of keypoints is a limitation of our present approach, we can also observe the capabilities of our novel descriptor learning method. Our method produces similar metrics to the state of the art, as reflected in the Table \ref{tab:drid}. The maximum difference in this case is around a 3\% on IoM and the minimum is 0.003 in LPIPS and around 1\% in both DICE and SM. However, due to the lower number of registered images (due to the aforementioned lack of keypoints), the error metrics are notably increased in Table \ref{tab:drid_2} which takes into account the number of registered images.


On the other hand, LongDRS offers a different setting for registration. While DeepDRiD generally has medium to low overlapping (depending on how the images are captured), LongDRS has images with extremely low overlapping as well as high overlapping, with a range of in-between values. Images such as the one in Figure \ref{fig:ldrs_a} are common. These images do not provide much possibility for visual analysis due to their low overlap, which is why we have only included one example in the Figure, even if they are very abundant in this dataset. In this sense, our method is, generally, either able to align them correctly or it fails completely due to the lack of keypoints. Moreover, this dataset also provides two visits, adding the extra complication brought by the passage of time and possible disease progression or the appearance of new lesions. Regarding the results of our approaches in this dataset, we can state that the results are satisfactory. Disregarding the number of registered pairs, as shown in Table \ref{tab:ldrs}, our method obtains results very close to SuperRetina. Moreover, even with the high penalization derived from the extremely low overlapping images (captured for image mosaicking not to be registered among themselves) which is enhanced by the low amount of detected keypoints by our method, we still manage to show competitive performance, as demonstrated by Table \ref{tab:ldrs_2}.  

The two separate visits on the LongDRS dataset allow to analyze the registrations in an intra and inter-visit manner. As previously mentioned, the intra-visit registrations have lower overlapping while the inter-visit one may have higher overlapping but can also have differences due to the one year gap in capture times. In this regard, we can observe in Tables \ref{tab:ldrs_inter} and \ref{tab:ldrs_intra} that our approach obtains significantly better results on the inter-visit registration than on the intra-visit registration. This is can also be seen with SuperRetina. More importantly, the gap presented by our approach to SuperRetina is lower in the inter-visit registration. These results are in-line with the previous explanations, as our method is heavily penalized in low overlapping scenarios. Therefore, in the inter-visit evaluation, where there are proportionally less low-overlapping pairs, our method closes the gap to SuperRetina which uses more keypoints. Consequently, in this evaluation our method also registers a higher proportion of the total of images.

Overall, our approach provided satisfactory results, matching state of the art performance in FIRE as well as the two novel datasets, DeepDRiD and LongDRS. These two datasets have significant differences from one to another as well as to FIRE. Therefore, performing accurately on all of them demonstrates that our method is able to generalize CF registration beyond the standard benchmark dataset FIRE.


\section{Conclusions and future work}

In this paper we study the effect of several losses in a state of the art color fundus registration framework. Our framework employs supervised natural keypoints from the retina, blood vessel crossovers and bifurcations. Using these keypoints we propose to learn their corresponding descriptors using a self-supervised contrastive learning approach. For this, we create a multi-viewed batch consisting of augmentations of the original dataset images. This way, each keypoint will have multiple positives and negatives which helps in improving the learning process. Using this framework we test four different losses: SupCon Loss, N-Pair, Info-NCE and FastAP, adapting them to the multi-positive multi-negative framework (MP-N-Pair, MP-InfoNCE) so that the results are all directly comparable. 

We train this method using the public DRIVE dataset and test it using the standard benchmark dataset FIRE. Furthermore we also test our method using two novel public datasets that have never been used for this purpose, for which we are releasing the image paring files. In this regard, the FastAP loss provided the best results across all the tests and metrics in the three datasets used. Furthermore, FastAP produces results competitive with the best deep learning state of the art despite our method being significantly more straightforward. However, our approach is limited by the amount of detected keypoints. As we use crossovers and bifurcations, the number of detected keypoints is limited by the morphology of the retina of each eye and each individual. Therefore, in cases of very low overlapping our method can fail due to lack of keypoints. In future work it is our intention to augment our current set of keypoints with other relevant ones, like other blood vessel keypoints.







\section*{Acknowledgments}

This work is supported by Ministerio de Ciencia e Innovación, Government of Spain, through the \mbox{PID2019-108435RB-I00}, \mbox{TED2021-131201B-I00}, and \mbox{PDC2022-133132-I00} 
research projects; Consellería  de  Cultura,  Educación e Universidade, Xunta de Galicia, through Grupos de Referencia Competitiva ref. \mbox{ED431C 2020/24}, predoctoral fellowship ref. \mbox{ED481A 2021/147} and the postdoctoral fellowship ref. \mbox{ED481B-2022-025}.

\bibliographystyle{unsrt}
\bibliography{references}  






\end{document}